\documentclass[acmtog]{acmart}
\acmSubmissionID{189}

\usepackage{booktabs} 
\usepackage{graphicx}
\usepackage{amsmath}
\usepackage{booktabs}
\usepackage{multirow}
\usepackage{colortbl}
\usepackage{makecell}
\usepackage{diagbox}
\usepackage{hyperref}

\usepackage[ruled]{algorithm2e} 

\SetAlFnt{\small}
\SetAlCapFnt{\small}
\SetAlCapNameFnt{\small}
\SetAlCapHSkip{0pt}

\copyrightyear{2024} 
\acmYear{2024} 
\setcopyright{acmlicensed}\acmConference[SA Conference Papers '24]{SIGGRAPH Asia 2024 Conference Papers}{December 3--6, 2024}{Tokyo, Japan}
\acmBooktitle{SIGGRAPH Asia 2024 Conference Papers (SA Conference Papers '24), December 3--6, 2024, Tokyo, Japan}
\acmDOI{10.1145/3680528.3687565}
\acmISBN{979-8-4007-1131-2/24/12}

\begin{document}
\newcommand\methodname{GVHMR }

\newcommand{\figref}[1]{Fig.~\ref{#1}}
\newcommand{\tabref}[1]{Tab.~\ref{#1}}
\newcommand{\secref}[1]{Sec.~\ref{#1}}
\newcommand{\AlgRef}[1]{Algorithm~\ref{#1}}
\newcommand{\equref}[1]{Eq.~\ref{#1}}

\newcommand{\tocite}[1]{\textcolor{red}{[TOCITE]}}

\newcommand{\sd}[2]{\textcolor{cyan}{[Sida: (#1) \textcolor{gray}{#2}]}}
\newcommand{\szh}[2]{\textcolor{green}{[Zehong: (#1) \textcolor{gray}{#2}]}}
\newcommand{\rh}[2]{\textcolor{blue}{[Ruizhen: (#1) \textcolor{blue}{#2}]}}
\newcommand{\0}{\phantom{0}}

\newcommand{\notice}[1]{\textcolor{cyan}{#1}}
\title{World-Grounded Human Motion Recovery via Gravity-View Coordinates
}

\author{Zehong Shen}
\authornote{Equal contribution.}
\email{hz_szh@zju.edu.cn}
\affiliation{%
 \institution{State Key Laboratory of CAD\&CG, Zhejiang University}
 \country{China}
}

\author{Huaijin Pi}
\authornotemark[1]
\email{hz_szh@zju.edu.cn}
\affiliation{%
 \institution{The University of Hong Kong}
 \country{China}
}

\author{Yan Xia}
\affiliation{%
 \institution{State Key Laboratory of CAD\&CG, Zhejiang University}
 \country{China}
}

\author{Zhi Cen}
\affiliation{%
 \institution{State Key Laboratory of CAD\&CG, Zhejiang University}
 \country{China}
}

\author{Sida Peng}
\authornote{Corresponding author.}
\affiliation{%
 \institution{Zhejiang University}
 \country{China}
}

\author{Zechen Hu}
\affiliation{%
 \institution{Deep Glint}
 \country{China}
}

\author{Hujun Bao}
\affiliation{%
 \institution{State Key Laboratory of CAD\&CG, Zhejiang University}
 \country{China}
}

\author{Ruizhen Hu}
\affiliation{%
 \institution{Shenzhen University}
 \country{China}
}

\author{Xiaowei Zhou}
\email{xwzhou@zju.edu.cn}
\affiliation{%
 \institution{State Key Laboratory of CAD\&CG, Zhejiang University}
 \country{China}
}

\begin{abstract}

    We present a novel method for recovering world-grounded human motion from monocular video.
    The main challenge lies in the ambiguity of defining the world coordinate system, which varies between sequences. 
    Previous approaches attempt to alleviate this issue by predicting relative motion in an autoregressive manner, but are prone to accumulating errors. 
    Instead, we propose estimating human poses in a novel Gravity-View (GV) coordinate system, which is defined by the world gravity and the camera view direction.
    The proposed GV system is naturally gravity-aligned and uniquely defined for each video frame, largely reducing the ambiguity of learning image-pose mapping. 
    The estimated poses can be transformed back to the world coordinate system using camera rotations, forming a global motion sequence.
    Additionally, the per-frame estimation avoids error accumulation in the autoregressive methods.
    Experiments on in-the-wild benchmarks demonstrate that our method recovers more realistic motion in both the camera space and world-grounded settings, outperforming state-of-the-art methods in both accuracy and speed.
    The code is available at \url{https://zju3dv.github.io/gvhmr}.

\end{abstract}



\begin{teaserfigure}
    \centering
    \includegraphics[width=1.0\textwidth]{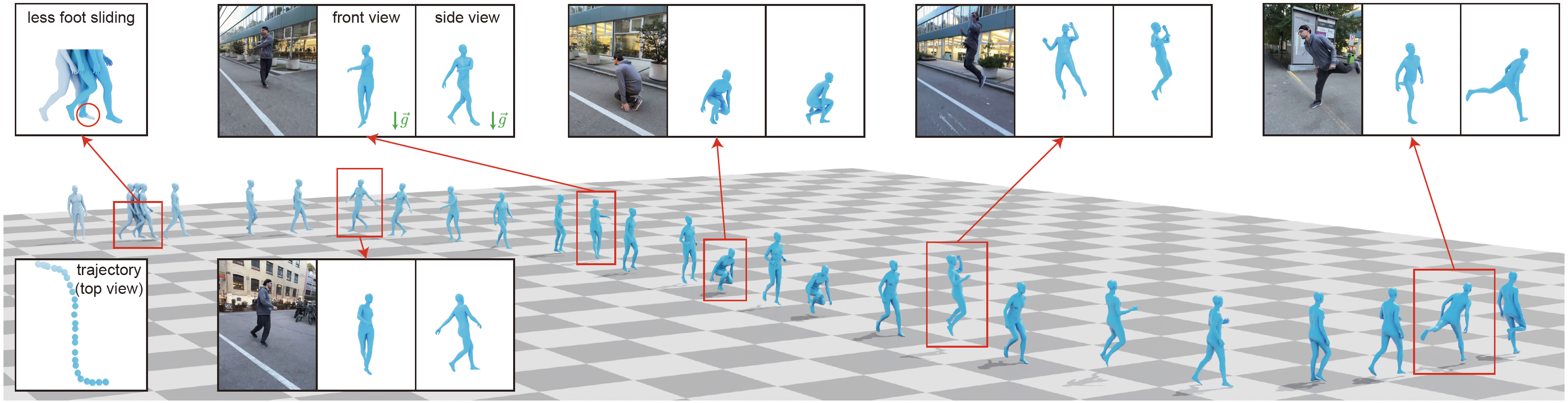}
    \caption{\textbf{Overview.}
        Given an in-the-wild monocular video, our method accurately regresses World-Grounded Human Motion:
        \textbf{4D human poses and shapes in a gravity-aware world coordinate system}.
        The proposed network, excluding preprocessing (2D human tracking, feature extraction, relative camera rotation estimation), takes 280 ms to process a 1430-frame video (approximately 45 seconds) on an RTX 4090 GPU.
    }
    \label{fig:teaser}
\end{teaserfigure}

\maketitle
\section{Introduction}

World-Grounded Human Motion Recovery (HMR) aims to reconstruct continuous 3D human motion within a gravity-aware world coordinate system.
Unlike conventional motion captured in the camera frame~\cite{hmr}, world-grounded motion is inherently suitable as foundational data for generative and physical models, such as text-to-motion generation~\cite{t2m,mdm} and humanoid robot imitation learning~\cite{human2humanoid}.
In these applications, motion sequences must be high-quality and consistent in a gravity-aware world coordinate system.

Most existing HMR methods can recover promising camera-space human motion from videos~\cite{vibe,mpsnet,glot}.
To recover the global motion, a straightforward approach is to use camera poses~\cite{dpvo} to transform camera-space motion to world-space.
However, the results are not guaranteed to be gravity-aligned, and errors in translations and poses can accumulate over time, resulting in implausible global motion.
Recent work, WHAM~\cite{wham}, attempts to recover global motion by autoregressively predicting relative global poses with RNN.
While this method achieves significant improvements, it requires a good initialization and suffers from accumulated errors over long sequences, making it challenging to maintain consistency in the gravity direction.
We believe the inherent challenge stems from the ambiguity in defining the world coordinate system.
Given the world coordinate axes, any rotation around the gravity axis defines a valid gravity-aware world coordinate system.

\begin{figure}
    \centering
    \includegraphics[width=0.9\columnwidth]{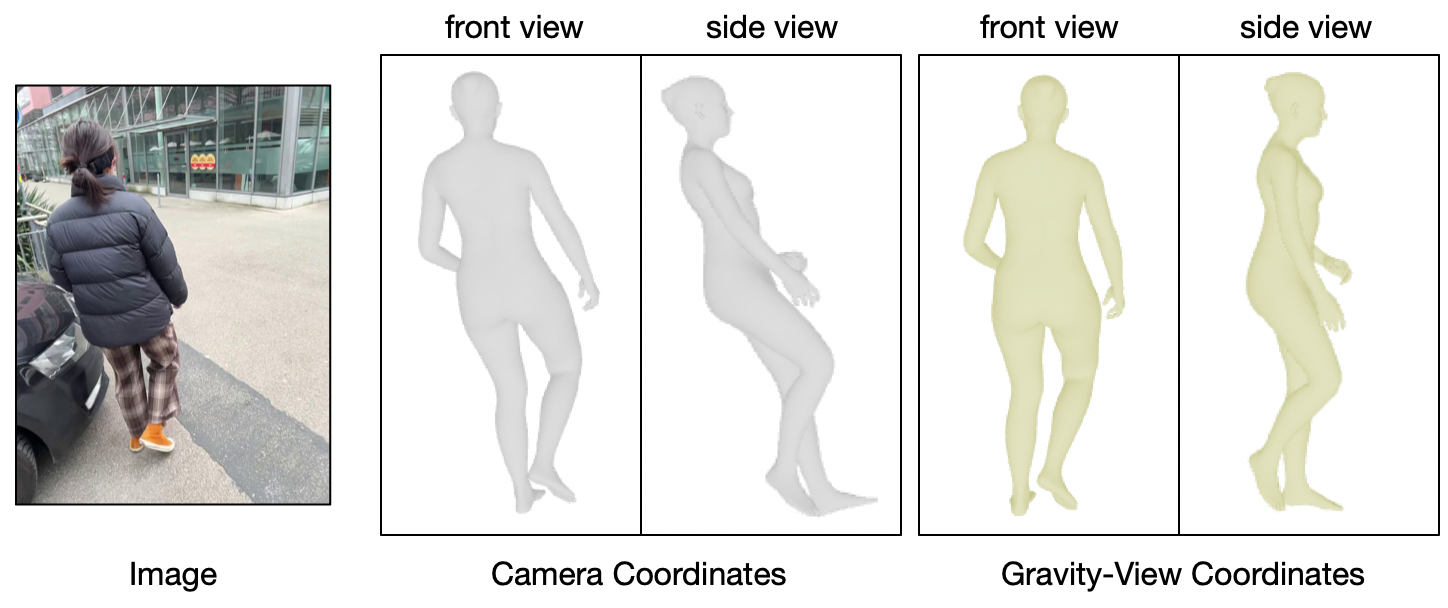}
    \vspace{-2mm}
    \caption{
        \textbf{Comparison of coordinate systems}. 
        In camera coordinates, a person may appear inclined due to the camera's roll and pitch movement.
        In contrast, in GV coordinates, the person is naturally aligned with gravity.
        }
    \label{fig:gv_over_c}
\end{figure}

In this work, we propose \methodname to estimate gravity-aware human poses for each frame and then compose them with gravity constraints to avoid accumulated errors in the gravity direction.
This design is motivated by the observation that, for a person in any image, we humans are able to easily infer the gravity-aware human pose, as shown in ~\figref{fig:gv_over_c}.
Additionally, given two consecutive frames, it is intuitively easier to estimate the 1-degree-of-freedom rotation around the gravity direction, compared to the full 3-degree-of-freedom rotation.
Therefore, we propose a novel Gravity-View (GV) coordinate system, defined by the gravity and camera view directions.
Using the GV system, we develop a network that predicts the gravity-aware human orientation.
We also propose a recovery algorithm to estimate the relative rotation between GV systems, enabling us to align all frames into a consistent gravity-aware world coordinate system.



Thanks to the GV coordinates, we can process human rotations in parallel over time.
We propose a transformer~\cite{transformer} model enhanced with Rotary Positional Embedding (RoPE)~\cite{rope} to directly regress the entire motion sequence.
Compared to the commonly used absolute position encoding, RoPE better captures the relative relationships between video frames and handles long sequences more effectively.
During inference, we introduce a mask to limit each frame's receptive field, avoiding the complex sliding windows and enabling parallel inference for infinitely long sequences.
Additionally, we predict stationary labels for hands and feet, which are used to refine foot sliding and global trajectories.

In summary, our contributions are threefold: 
1. We propose a novel Gravity-View coordinate system and the global orientation recovery method to reduce the cumulative errors in the gravity direction.
2. We develop a Transformer model enhanced by RoPE to generalize to long sequences and improve motion estimation.
3. We demonstrate the effectiveness of our approach through extensive experiments, showing that it outperforms previous methods in both in-camera and world-grounded accuracy.

\section{Related Works}

\paragraph{Camera-Space Human Motion Recovery}
Recent studies in 3D human recovery predominantly use parametric human models such as SMPL~\cite{smpl, smplx}.
Given a single image or video, the target is to align the human mesh precisely with the 2D images.
Early methods~\cite{smplx,smplify} employ optimization-based approaches by minimizing the reprojection error.
Recently, regression-based methods~\cite{hmr,hmr2} trained on a large amount of data predict the SMPL parameters from the input image directly.
Many efforts have been made to improve the accuracy by specialized design architectures~\cite{pymafx,niki}, part-based reasoning~\cite{pare,hybrik}, and incorporating camera parameters~\cite{cliff, spec}.
HMR2.0~\cite{hmr2} designs a ViT architecture~\cite{transformer} and outperforms the previous methods.
To utilize temporal cues, ~\cite{shi2020motionet} uses deep networks to predict skeleton pose sequence directly from videos.
To recover the human mesh, most methods build upon the HMR pipeline.
~\cite{hmmr} adopts a convolutional encoder.
~\cite{vibe,meva,tcmr} apply RNN successfully.
~\cite{dsd-satn} introduces self-attention to CNN.
~\cite{maed,glot} employ a transformer encoder to extract temporal information.

Although these methods can accurately estimate human pose, their predictions are all in the camera-space. Consequently, when the camera moves, the human motion becomes physically implausible.

\paragraph{World-Grounded Human Motion Recovery}
Traditionally, estimating human motion in a gravity-aware world coordinate system requires additional floor plane calibration or gravity sensors.
In multi camera capture systems~\cite{rich,h36m}, calibration boards are placed on the ground to reconstruct the ground plane and global scale. 
IMU-based methods~\cite{3dpw,emdb,transpose} use gyroscopes and accelerometers to estimate the gravity direction and then project human motion onto the gravity direction.
Recently, researchers put efforts to estimate global human motion from a monocular video.
~\cite{humandynamic} reconstructs human motion using physics law but requires a provided scene.
Methods like~\cite{glamr,dandd} predicts the global trajectory from locomotion cues.
However, the camera motion and human motion are coupled, which make the results noisy.
SLAHMR~\cite{slahmr} and PACE~\cite{pace} further integrate SLAM~\cite{droid, dpvo} and pre-learned human motion priors~\cite{humor} in an optimization framework.
Although these methods achieve promising results, the optimization process is time-consuming and faces convergence issues with long video sequences.
Furthermore, these methods do not obtain gravity-aligned human motion.

\begin{figure*}[t]
    \centering
    \includegraphics[width=0.88\linewidth]{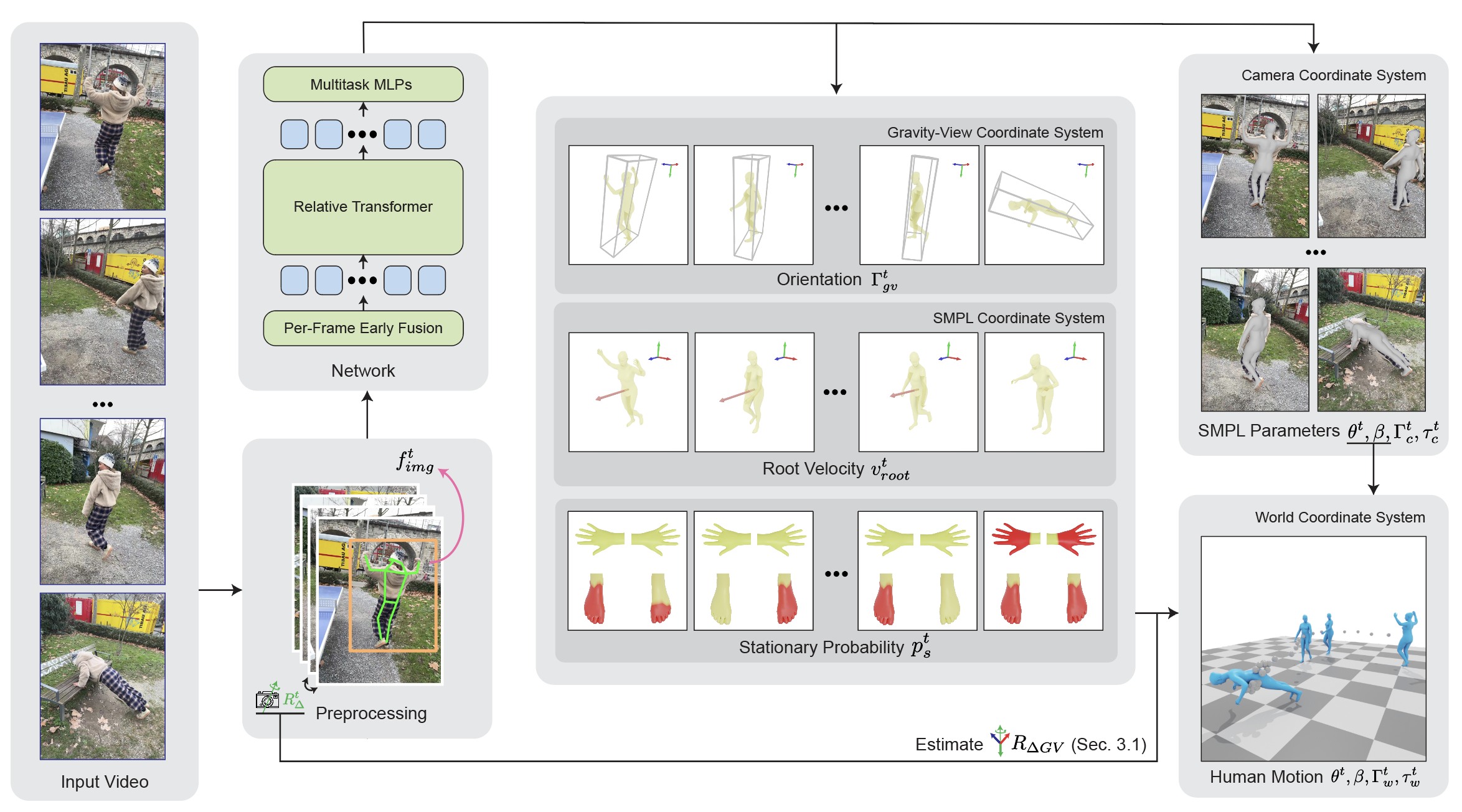}
    \vspace{-2mm}
    \caption{\textbf{Overview of the proposed framework.}
        Given a monocular video (left), following WHAM~\cite{wham}, \methodname preprocesses the video by tracking the human bounding box, detecting 2D keypoints, extracting image features, and estimating camera relative rotation using visual odometry or a gyroscope.
        \methodname then fuses these features into per-frame tokens, which are processed with a relative transformer and multitask MLPs. The outputs include: (1) intermediate representations (middle), i.e. human orientation in the Gravity-View coordinate system, root velocity in the SMPL coordinate system, and the stationary probability for predefined joints; and (2) camera frame SMPL parameters (right-top).
        Finally, the global trajectory (right-bottom) is recovered by transforming the intermediate representations to the world coordinate system, as described in Sec. 3.1.
    }
    \label{fig:pipeline}
\end{figure*}

The most relevant work is WHAM~\cite{wham}, which directly regresses per-frame pose and translation in an autoregressive manner.
However, their method relies on a good initialization and the performance drops in long-term motion recovery due to error accumulation.
Two concurrent works also focus on world-grounded human motion recovery.
WHAC~\cite{whac} uses visual odometry~\cite{dpvo} to transform camera coordinate results to a world coordinate system and relies on another network to refine global trajectory.
TRAM~\cite{tram} employs SLAM~\cite{droid} to recover camera motion and uses the scene background to derive the motion scale.
They also transform the camera coordinate results into a world coordinate system.
In contrast to their methods, \methodname does not require additional refinement networks and can directly predict the world-grounded human motion.

\section{Method}
\label{sec:method}

Given a monocular video $\{I^{t}\}_{t=0}^{T}$,
we formulate the task as predicting:
(1) the local body poses $\{\theta^{t}\in \mathbb{R} ^{21\times 3}\}_{t=0}^{T}$ and shape coefficients $\beta\in \mathbb{R}^{10}$ of SMPL-X,
(2) the human trajectory from SMPL space to the camera space, including the orientation $\{\Gamma_c^{t}\in \mathbb{R}^{3}\}_{t=0}^{T}$ and translation $\{\tau_c^{t}\in \mathbb{R}^{3}\}_{t=0}^{T}$,
(3) the trajectory to the world space, including the orientation $\{\Gamma_w^{t}\in \mathbb{R}^{3}\}_{t=0}^{T}$ and translation $\{\tau_w^{t}\in \mathbb{R}^{3}\}_{t=0}^{T}$.

An overview of the proposed pipeline is shown in \figref{fig:pipeline}.
In \secref{sec:global_traj}, we first introduce the global trajectory representation and discuss its advantages over previous trajectory representations.
Then, \secref{sec:net_arch} describes a specially designed network architecture as well as post-process techniques for predicting the targets.
Finally, implementation details are presented in \secref{sec:impl}.

\subsection{Global Trajectory Representation}
\label{sec:global_traj}














Global human trajectory $\{\Gamma^t_w, \tau^t_w\}$ refers to the transformation from SMPL space to the {\it gravity-aware world space} $W$.
However, the definition of $W$ varies, as any rotation of $W$ around the gravity direction is valid, leading to different $\Gamma_w$ and $\tau_w$.
We propose to first recover a gravity-aware human pose for each image, then transform these poses to a consistent global trajectory.
This approach is inspired by the observation that humans can easily infer the orientation and gravity direction of a person in an image.
And for consecutive frames, estimating the relative rotation around the gravity direction is intuitively easier and more robust.

Specifically, for each image, we use the world gravity direction and the camera's view direction (i.e., the normal vector of the image plane) to define Gravity-View (GV) Coordinates.
The proposed new GV coordinate system is mainly used to resolve the rotation ambiguity, so we only predict the per-frame human orientation $\Gamma^{t}_{GV}$ relative to the GV system.
When the camera moves, we compute the relative rotation between the GV systems of two adjacent frames with relative camera rotations $R^t_{\Delta}$, thus transforming all $\Gamma^{t}_{GV}$ to a consistent gravity-aware global space.
For global translation, following~\cite{humor,wham}, we predict the human displacement in the SMPL coordinate system from time $t$ to $t+1$, and finally roll out in the aforementioned world reference frame.






\paragraph{Gravity-View Coordinate System}
\begin{figure}
    \centering
    \includegraphics[width=\columnwidth]{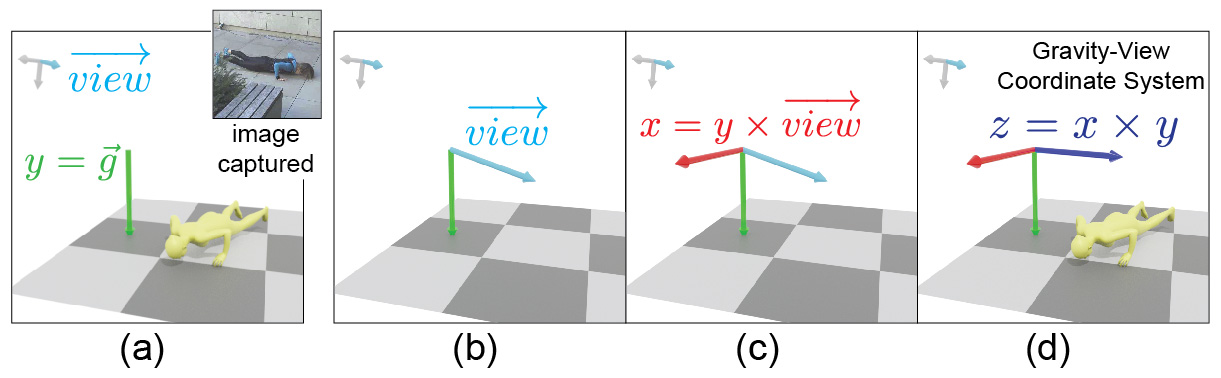}
    \vspace{-6mm}
    \caption{
        \textbf{Gravity-View (GV) coordinate system}, defined by the gravity direction and the camera view direction.
        (Refer to Sec.~3.1 for details).
        }
    \label{fig:cg_coord}
\end{figure}
As illustrated in \figref{fig:cg_coord}, (a) given a person with orientation $\Gamma_c$ and a gravity direction $\vec{g}$ both described in the camera space:
(b) the y-axis of the GV coordinate system aligns with the gravity direction $\vec{g}$, i.e., $\vec{y} = \vec{g}$;
(c) the x-axis is perpendicular to both the camera view direction $\overrightarrow{view}=[0,0,1]^{T}$ and $\vec{y}$ by cross-product, i.e., $\vec{x} = \vec{y} \times  \overrightarrow{view} $;
(d) finally, the z-axis is calculated by the right-hand rule, i.e., $\vec{z} = \vec{x} \times \vec{y}$.
After obtaining these axes, we can re-calculate the person's orientation in the GV coordinate system as our learning target: $\Gamma_{GV}= R_{c2GV} \cdot \Gamma_c = [\vec{x}, \vec{y}, \vec{z}]^T \cdot \Gamma_c$.

\paragraph{Recovering Global Trajectory}
\begin{figure}
    \centering
    \includegraphics[width=0.65\columnwidth]{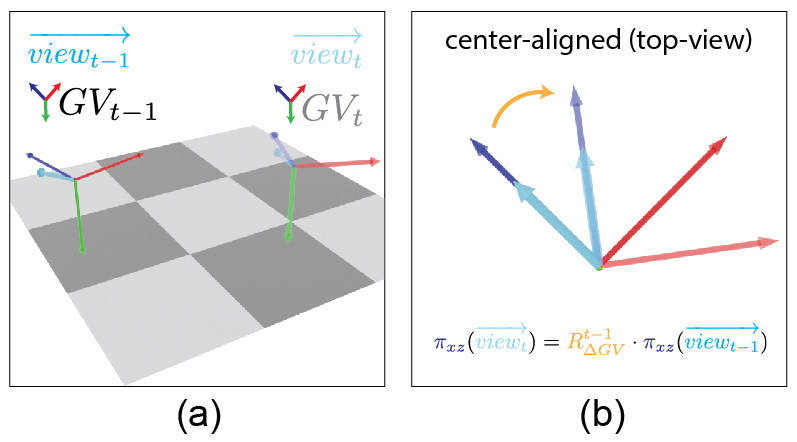}
    \vspace{-3mm}
    \caption{
        \textbf{Relative rotation between two GV coordinate systems.}
        (a) shows two adjacent GV coordinate systems and the \textcolor{cyan}{camera view directions}.
        (b) illustrates the relative rotation between two GV systems. $R_{\Delta GV}$ occurs \textbf{exclusively} around the y-axis (gravity direction).
    }
    \label{fig:compute_rel_R_GV}
    \vspace{-2mm}
\end{figure}
It is noteworthy that an independent $GV_t$ exists for each input frame $t$, where we predict the person's orientation ${\Gamma_{GV}^{t}}$.
To recover a consistent global trajectory $\{\Gamma_w^{t},\tau_w^{t}\}$, 
all orientations must be transformed to a common reference system.
In practice, we use $GV_0$ as the world reference system $W$.

To begin with, in the special case of a static camera, the $GV_t$ systems are identical across all frames.
Therefore, the human global orientation $\{\Gamma_w^t\}$ is equivalent to $\{\Gamma_{GV}^t\}$.
The translation $\{\tau_w^t\}$ is obtained by transforming all predicted local velocities ${v_{root}}$ into the world coordinate system using the orientations $\{\Gamma_w^t\}$ and then performing a cumulative sum:
\begin{align}
    \label{eq:translation}
    \tau_w^t =\begin{cases}
        [0,0,0]^T,                                      & t=0, \\
        \sum_{i=0}^{t-1} \Gamma_{w}^{i} v_{root}^{i}, & t>0.
    \end{cases}
\end{align}

For a moving camera,
we first compute the rotation $R_{\Delta GV}^{t}$ between the GV coordinate systems of frame $t$ to frame $t-1$ by leveraging the input camera relative rotations $R_{\Delta}^t$, the predicted human orientations $\Gamma_c^t$ and $\Gamma_{GV}^t$.
As illustrated in \figref{fig:compute_rel_R_GV}, we first calculate the rotation from camera to GV coordinate system at frame $t$: $R_{c2gv}^{t} = \Gamma_{GV}^{t} \cdot (\Gamma_c^t)^{-1}$.
Then, the camera view direction $\overrightarrow{view}^{t}_{c} = [0,0,1]^T$ is transformed to the GV coordinate system as $\overrightarrow{view}^{t}_{GV} = R_{c2gv}^{t} \cdot \overrightarrow{view}^{t}_c$.
We use the camera's relative transformation to rotate this view direction to frame $t-1$, i.e., $\overrightarrow{view}^{t-1} = (R_{\Delta}^{t})^{-1} \cdot \overrightarrow{view}^{t}$.
Since the rotation between the $GV_t$ systems is always around the gravity vector, we can calculate the rotation matrix $R_{\Delta GV}^{t}$ by projecting the view directions $\overrightarrow{view}^{t-1}$ and $\overrightarrow{view}^{t}$ onto the xz-plane and computing the angle between them.
After obtaining $\{R_{\Delta GV}^{t}\}$ of the entire input sequences, we can roll out to the first frame's GV coordinate system for all frames:
\begin{align}
    \Gamma_w^t =\begin{cases}
        \Gamma_{GV}^0,                                         & t=0, \\
        \prod_{i=1}^{t} R_{\Delta GV}^{i} \cdot \Gamma_{GV}^t, & t>0.
    \end{cases}
\end{align}
This formulation also applies to static cameras, as the transformation $R_{\Delta GV}^{t}$ is the identity transformation in this case.
Finally, the translation is obtained using the same method as described in \equref{eq:translation}.

The human orientation in the GV coordinate system is well-suited for deep network learning, given that the establishment of the GV coordinate system is determined from the input images.
It also ensures that the learned global orientation is naturally gravity-aware.
We have also found this approach beneficial for learning local pose and shape, as demonstrated in the ablation study~\tabref{tab:ablation_rich}.
In the rotation recovery algorithm between GV systems, we utilize the consistency of the y-axis in the GV system to systematically avoid cumulative errors in the gravity direction.
This also mitigates potential errors in camera rotation estimation, resulting in our method achieving similar results under both GT Gyro and DPVO estimated relative camera rotations, as shown in ~\tabref{tab:quant_global}.
Compared to WHAM, our method does not require initialization and can predict in parallel without the need for autoregressive prediction.

\subsection{Network Design}
\label{sec:net_arch}

\begin{figure}
    \centering
    \includegraphics[width=0.6\columnwidth]{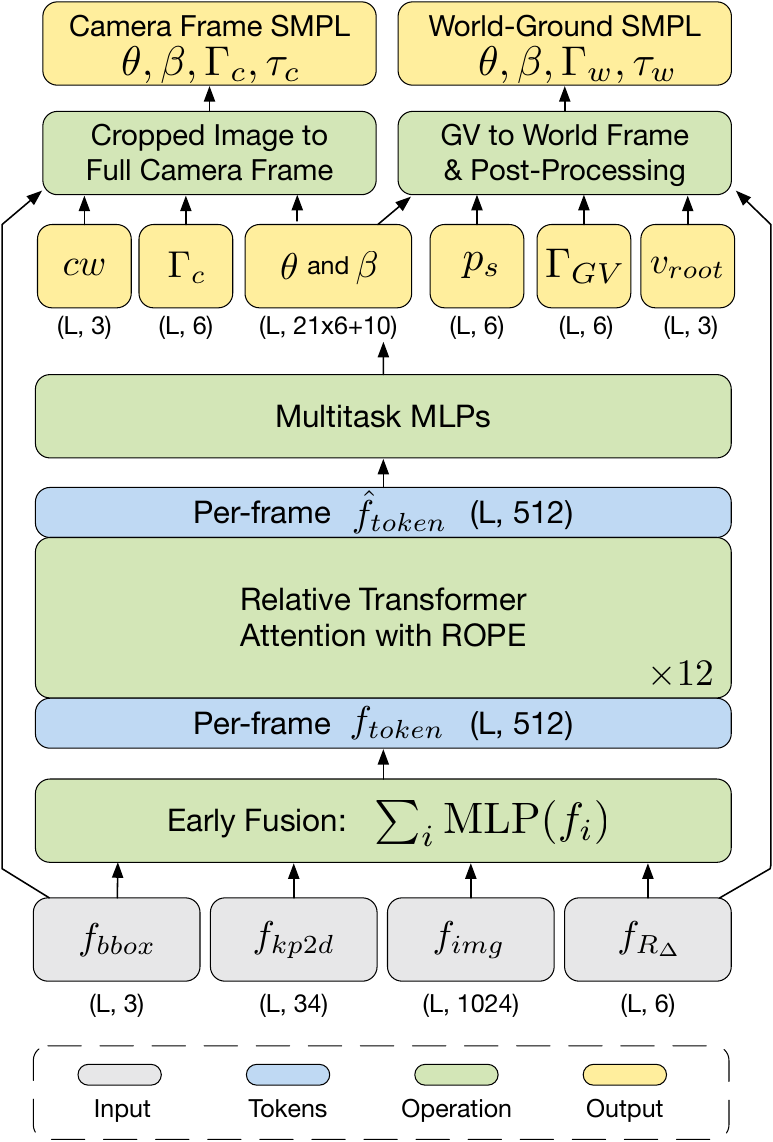}
    \vspace{-2mm}
    \caption{ \textbf{Network architecture.}
        The input features are fused into per-frame tokens by the early-fusion module, processed by the relative transformer, and then output by multitask MLPs as intermediate representations.
        The weak-camera parameter $cw$ is restored to the camera frame $\tau_c$ following~\cite{cliff}. 
        The predicted $\Gamma_{GV}$ and $v_{root}$ are converted to the world frame $\Gamma_w$ and $\tau_w$, as described in \secref{sec:global_traj}.
        Finally, we use joint stationary probabilities $p_s$ to post-process the global motion.
    }
    \vspace{-2mm}
    \label{fig:net_arch}
\end{figure}

\paragraph{Input and preprocessing}
The network design is shown in \figref{fig:net_arch}.
Inspired by WHAM~\cite{wham}, we first preprocess the input video into four types of features: bounding boxes\cite{yolov8,cliff}, 2D keypoints~\cite{vitpose}, image features~\cite{hmr2}, and relative camera rotations~\cite{dpvo}.
Then, in the early-fusion module, we use individual MLPs to map these features to the same dimension.
These vectors are then element-wise added to obtain per-frame tokens $\{f_{\text{token}}^t \in \mathbb{R}^{512}\}$.
These tokens are processed by a Relative Transformer, where we introduce rotary positional encoding (RoPE)~\cite{rope} to enable the network to focus on relative position features.
Additionally, we implement a receptive-field-limited attention mask to improve the network’s generalization ability when testing on long sequences.

\paragraph{Rotary positional embedding.}
Absolute positional embedding is a common approach for transformer architectures in human motion modeling.
However, this implicitly reduces the model's ability to generalize to long sequences because the model is not trained on positional encodings beyond the training length.
We argue that the absolute position of human motions is ambiguous (e.g., the start of a motion sequence can be arbitrary).
In contrast, the relative position is well-defined and can be easily learned.

Here we introduce rotary positional embedding to inject relative features into temporal tokens, where the output $\mathbf{o}^t$ of the $t$-th token after the self-attention layer is calculated via:
\begin{align}
    \mathbf{o}^t = \sum_{i \in T} \underset{s \in T}{\operatorname{Softmax}}\left(a^{ts}\right)^i \mathbf{W}_v f_{token}^i \\
    a^{ts} = (\mathbf{W}_q f_{token}^t)^{\top} \mathbf{R}\left(\mathbf{p}^s - \mathbf{p}^t\right) (\mathbf{W}_k f_{token}^s)
\end{align}
where $\mathbf{W_q}$, $\mathbf{W_k}$, $\mathbf{W_v}$ are the projection matrix, $\mathbf{R}(\cdot) \in \mathbb{R}^{512 \times 512}$ is the rotary encoding of the relative position between two tokens, and $\mathbf{p}^t$ indicates the temporal index of the $t$-th token.
Following the definition in RoPE, we divide the 512-dimensional space into 256 subspaces and combine them using the linearity of the inner product. $\mathbf{R}(\cdot)$ is defined as:
\begin{equation}
    \mathbf{R}(\mathbf{p})=\left(\begin{array}{ccc}
            \hat{\mathbf{R}}\left(\alpha_1^{\top} \mathbf{p}\right) &        & \mathbf{0}                                                  \\
                                                                    & \ddots &                                                             \\
            \mathbf{0}                                              &        & \hat{\mathbf{R}}\left(\alpha_{256}^{\top} \mathbf{p}\right)
        \end{array}\right), \hat{\mathbf{R}}(\theta)=\left(\begin{array}{cc}
            \cos \theta & -\sin \theta \\
            \sin \theta & \cos \theta
        \end{array}\right) ,
\end{equation}
where $\alpha_i$ is pre-defined frequency parameters.

At inference time, we further introduce an attention mask~\cite{alibi} and the self-attention becomes:
\begin{align}
    \mathbf{o}^t = \sum_{i \in T} \underset{s \in T}{\operatorname{Softmax}}\left(a^{ts} + m^{ts}\right)^i \mathbf{W}_v f_{token}^i \\
    m^{ts} = \begin{cases}
        0,       & \text{if } -L < t - s < L, \\
        -\infty, & \text{otherwise}.
    \end{cases}
\end{align}
where $L$ is the maximum training length.
The token $t$ attends only to tokens within $L$ relative positions.
Consequently, the model can generalize to arbitrarily long sequences without needing autoregressive inference techniques, such as sliding-window.

\paragraph{Network outputs.}
After the relative transformer, the $f_{\text{token}}'$ are processed by multitask MLPs to predict multiple targets, including the weak-perspective camera parameters $cw$, the human orientation in the camera frame $\Gamma_c$, the SMPL local pose $\theta$, the SMPL shape $\beta$, the stationary label $p_j$, the global trajectory representation $\Gamma_{GV}$ and $v_{root}$.
To get the camera-frame human motion, we follow the standard CLIFF~\cite{cliff} to transform the weak-perspective camera to full-perspective.
For the world-grounded human motion, we recover the global trajectory as described in \secref{sec:global_traj}.


\paragraph{Post-processing}
The proposed network learns smooth and realistic global movement from the training data.
Inspired by WHAM, we additionally predict joint stationary probabilities to further refine the global motion.
Specifically, we predict the stationary probabilities for the hands, toes, and heels, and then update the global translation frame-by-frame to ensure that the static joints remain at fixed points in space as much as possible.
After updating the global translation, we calculate the fine-grained stationary positions for each joint (see the algorithm in the supplementary).
These target joint positions are then passed into an inverse kinematics process to solve the local poses, mitigating physically implausible effects like foot-sliding.
We use a CCD-based IK solver~\cite{ccdik} with an efficient implementation~\cite{nsm}.



\paragraph{Losses}
We use the following losses for training:
Mean Squared Error (MSE) loss on predicted targets except for stationary probability, which uses Binary Cross-Entropy (BCE) loss.
Additionally, we use L2 loss on 3D joints, 2D joints, vertices, translation in the camera frame, and translation in the world coordinate system.
More details are provided in the supplementary material.

\subsection{Implementation details}
\label{sec:impl}
\methodname has 12 layers of transformer encoder.
Each attention unit has 8 heads.
The hidden dimension is 512.
The MLP has two linear layers with GELU activation.
\methodname is trained from scratch on a mixed dataset consisting of AMASS~\cite{amass}, BEDLAM~\cite{bedlam}, H36M~\cite{h36m}, and 3DPW~\cite{3dpw}.
During training, we augment the 2D keypoints following WHAM.
For AMASS, we simulate static and dynamic camera trajectories, generate bounding boxes, normalize the keypoints using these boxes from -1 to 1, and set image features to zero.
For other datasets that come with videos, we extract image features using a fixed encoder~\cite{hmr2}.
The training sequence length is set to $L=120$.
The model converges after 500 epochs with a batch size of 256. Training takes 13 hours on 2 RTX 4090 GPUs.

\newcolumntype{?}{!{\vrule width 0.75pt}}

\begin{table*}[tbh]
    \centering
    \caption{
    \textbf{World-grounded metrics.} We evaluate the global motion quality on the RICH~\cite{rich} and EMDB-2~\cite{emdb} dataset.
    Parenthesis denotes the number of joints used to compute WA-MPJPE$_{100}$, W-MPJPE$_{100}$ and Jitter.
    }
    \vspace{-4mm}
    \label{tab:quant_global}
    \setlength{\tabcolsep}{3pt}
    \renewcommand{\arraystretch}{1.2}
    \resizebox{0.80\textwidth}{!}
    {\small{
            \begin{tabular}{l?ccccc?ccccc}
                \cmidrule[0.75pt]{1-11}
                                                   & \multicolumn{5}{c}{RICH (24)} & \multicolumn{5}{c}{EMDB (24)}                                                                                                                                                                                                          \\
                \cmidrule(lr){2-6} \cmidrule(lr){7-11}

                Models                             & \scriptsize{WA-MPJPE$_{100}$} & \scriptsize{W-MPJPE$_{100}$}  & \scriptsize{RTE} & \scriptsize{Jitter} & \scriptsize{Foot-Sliding} & \scriptsize{WA-MPJPE$_{100}$} & \scriptsize{W-MPJPE$_{100}$} & \scriptsize{RTE} & \scriptsize{Jitter} & \scriptsize{Foot-Sliding} \\
                \cmidrule{1-11}
                DPVO\cite{dpvo} +HMR2.0\cite{hmr2} & 184.3                         & 338.3                         & \07.7            & 255.0               & 38.7                      & 647.8                         & 2231.4                       & 15.8             & 537.3               & 107.6                     \\
                GLAMR \cite{glamr}                 & 129.4                         & 236.2                         & \03.8            & \049.7              & 18.1                      & 280.8                         & 726.6                        & 11.4             & 46.3                & 20.7                      \\
                TRACE \cite{trace}                 & 238.1                         & 925.4                         & 610.4            & 1578.6              & 230.7                     & 529.0                         & 1702.3                       & 17.7             & 2987.6              & 370.7                     \\
                SLAHMR \cite{slahmr}               & \098.1                        & 186.4                         & 28.9             & \034.3              & \05.1                     & 326.9                         & 776.1                        & 10.2             & 31.3                & 14.5                      \\
                WHAM (w/ DPVO) \cite{wham}         & 109.9                         & 184.6                         & \04.1            & \019.7              & \03.3                       & 135.6                         & 354.8                        & \06.0            & 22.5                & \04.4                     \\
                WHAM (w/ GT gyro) \cite{wham}      & 109.9                         & 184.6                         & \04.1            & \019.7              & \03.3                     & 131.1                         & 335.3                        & \04.1            & 21.0                & \04.4                     \\
                \cmidrule{1-11}
                Ours (w/ DPVO)                     & \0\textbf{78.8}               & \textbf{126.3}                & \0\textbf{2.4}   & \0\textbf{12.8}     & \0\textbf{3.0}            & 111.0                         & 276.5                        & \02.0              & 16.7       & \0\textbf{3.5}            \\
                Ours (w/ GT gyro)                  & \0\textbf{78.8}               & \textbf{126.3}                & \0\textbf{2.4}   & \0\textbf{12.8}     & \0\textbf{3.0}            & \textbf{109.1}                & \textbf{274.9}               & \0\textbf{1.9}   & \textbf{16.5}       & \0\textbf{3.5}            \\                \cmidrule[0.75pt]{1-11}
            \end{tabular}
        }}

\end{table*}
\newcolumntype{?}{!{\vrule width 0.75pt}}

\begin{table*}[tbh]
    \centering
    \vspace{-2mm}
    \caption{
    \textbf{Camera-space metrics.} We evaluate the camera-space motion quality on the 3DPW \cite{3dpw}, RICH ~\cite{rich} and EMDB-1 ~\cite{emdb} datasets.
    $^*$ denotes models trained with the 3DPW training set.
    }

    \vspace{-4mm}
    \label{tab:quant_incam}
    \setlength{\tabcolsep}{3pt}
    \renewcommand{\arraystretch}{1.2}
    \resizebox{0.70\textwidth}{!}
    {\small{
            \begin{tabular}{cl?cccc?cccc?cccc}
                \cmidrule[0.75pt]{1-14}
                 &                           & \multicolumn{4}{c}{3DPW (14)} & \multicolumn{4}{c}{RICH (24)} & \multicolumn{4}{c}{EMDB (24)}                                                                                                                                                                                                \\
                \cmidrule(lr){3-6} \cmidrule(lr){7-10} \cmidrule(lr){11-14}

                 & Models                    & \scriptsize{PA-MPJPE}         & \scriptsize{MPJPE}            & \scriptsize{PVE}              & \scriptsize{Accel} & \scriptsize{PA-MPJPE} & \scriptsize{MPJPE} & \scriptsize{PVE} & \scriptsize{Accel} & \scriptsize{PA-MPJPE} & \scriptsize{MPJPE} & \scriptsize{PVE} & \scriptsize{Accel} \\
                \cmidrule{1-14}

                \multirow{6}{1em}{\rotatebox[origin=c]{90}{per-frame}}
                 & SPIN \cite{spin}          & 59.2                          & 96.9                          & 112.8                         & 31.4               & 69.7                  & 122.9              & 144.2            & 35.2               & 87.1                  & 140.3              & 174.9            & 41.3               \\
                 & PARE$^*$ \cite{pare}      & 46.5                          & 74.5                          & 88.6                          & --                 & 60.7                  & 109.2              & 123.5            & --                 & 72.2                  & 113.9              & 133.2            & --                 \\
                 & CLIFF$^*$ \cite{cliff}    & 43.0                          & 69.0                          & 81.2                          & 22.5               & 56.6                  & 102.6              & 115.0            & 22.4               & 68.1                  & 103.3              & 128.0            & 24.5               \\
                 & HybrIK$^*$ \cite{hybrik}  & 41.8                          & 71.6                          & 82.3                          & --                 & 56.4                  & 96.8               & 110.4            & --                 & 65.6                  & 103.0              & 122.2            & --                 \\
                 & HMR2.0 \cite{hmr2}        & 44.4                          & 69.8                          & 82.2                          & 18.1               & 48.1                  & 96.0               & 110.9            & 18.8               & 60.6                  & 98.0               & 120.3            & 19.8               \\
                 & ReFit$^*$ \cite{refit}    & 40.5                          & 65.3                          & 75.1                          & 18.5               & 47.9                  & 80.7               & 92.9             & 17.1               & 58.6                  & 88.0               & 104.5            & 20.7               \\
                \cmidrule{1-14}

                \multirow{12}{1em}{\rotatebox[origin=c]{90}{temporal}}
                 & TCMR$^*$ \cite{tcmr}      & 52.7                          & 86.5                          & 101.4                         & 6.0                & 65.6                  & 119.1              & 137.7            & 5.0                & 79.6                  & 127.6              & 147.9            & 5.3                \\
                 & VIBE$^*$ \cite{vibe}      & 51.9                          & 82.9                          & 98.4                          & 18.5               & 68.4                  & 120.5              & 140.2            & 21.8               & 81.4                  & 125.9              & 146.8            & 26.6               \\
                 & MPS-Net$^*$ \cite{mpsnet} & 52.1                          & 84.3                          & 99.0                          & 6.5                & 67.1                  & 118.2              & 136.7            & 5.8                & 81.3                  & 123.1              & 138.4            & 6.2                \\
                 & GLoT$^*$ \cite{glot}      & 50.6                          & 80.7                          & 96.4                          & 6.0                & 65.6                  & 114.3              & 132.7            & 5.2                & 78.8                  & 119.7              & 138.4            & 5.4                \\
                 & GLAMR \cite{glamr}        & 51.1                          & --                            & --                            & 8.0                & 79.9                  & --                 & --               & 107.7              & 73.5                  & 113.6              & 133.4            & 32.9               \\
                 & TRACE$^*$ \cite{trace}    & 50.9                          & 79.1                          & 95.4                          & 28.6               & --                    & --                 & --               & --                 & 70.9                  & 109.9              & 127.4            & 25.5               \\
                 & SLAHMR \cite{slahmr}      & 55.9                          & --                            & --                            & --                 & 52.5                  & --                 & --               & 9.4                & 69.5                  & 93.5               & 110.7            & 7.1                \\
                 & PACE \cite{pace}          & --                            & --                            & --                            & --                 & 49.3                  & --                 & --               & 8.8                & --                    & --                 & --               & --                 \\
                 & WHAM$^*$ \cite{wham}      & \textbf{35.9}                 & 57.8                          & 68.7                          & 6.6                & 44.3                  & 80.0               & 91.2             & 5.3                & 50.4                  & 79.7               & 94.4             & 5.3                \\
                \cmidrule{2-14}
                 & Ours$^*$                  & 36.2                          & \textbf{55.6}                 & \textbf{67.2}                 & \textbf{5.0}       & \textbf{39.5}         & \textbf{66.0}      & \textbf{74.4}    & \textbf{4.1}       & \textbf{42.7}         & \textbf{72.6}      & \textbf{84.2}    & \textbf{3.6}       \\
                \cmidrule[0.75pt]{1-14}
            \end{tabular}
        }}

\end{table*}

\section{Experiments}

\subsection{Datasets and Metrics}
\paragraph{Evaluation datasets.}
Following WHAM~\cite{wham}, we evaluate our method on three in-the-wild benchmarks: 3DPW~\cite{3dpw}, RICH~\cite{rich}, EMDB~\cite{emdb}.
We use RICH and EMDB-2 split to evaluate the global performance.
The RICH test set contains 191 videos captured with static cameras, totaling 59.1 minutes with accurate global human motion annotations.
The EMDB-2 is captured with moving cameras and contains 25 sequences totaling 24.0 minutes.
Additionally, we use RICH, EMDB-1 split, and 3DPW to evaluate the camera-coordinate performance.
EMDB-1 contains 17 sequences totaling 13.5 minutes, and 3DPW contains 37 sequences totaling 22.3 minutes.
We also test our method on internet videos for qualitative results (see supplementary video).

\paragraph{Metrics.}
We follow the evaluation protocol of~\cite{wham, slahmr}, using the code released by WHAM to 
apply FlipEval for test-time augmentation and evaluate our model's performance.
To compute world-coordinate metrics, we divide the predicted global sequences into shorter segments of 100 frames and align each segment to the ground-truth segment.
When the alignment is performed using the entire segment, we report the World-aligned Mean Per Joint Position Error (WA-MPJPE$_{100}$).
When the alignment is performed using the first two frames, we report the World MPJPE (W-MPJPE$_{100}$).
Additionally, to assess the error over the global motion, we evaluate the whole sequence for Root Translation Error (RTE, in $\%$), motion jittery (Jitter, in $m/s^3$), and foot sliding (FS, in $mm$).
The camera-coordinate metrics include the widely used MPJPE, Procrustes-aligned MPJPE (PA-MPJPE), Per Vertex Error (PVE), and Acceleration error (Accel, in $m/s^2$)~\cite{cliff, hmr2, slahmr, wham}.

\subsection{Comparison on Global Motion Recovery}
We compare our method with several state-of-the-art methods that recover global motion and a straightforward baseline method that combines the state-of-the-art camera-space method HMR2.0~\cite{hmr2} with a SLAM method (DPVO~\cite{dpvo}).
The GT gyro indicates the ground-truth camera rotation data in the EMDB dataset provided by the ARKit.
For the static camera in the RICH dataset, we set the camera transformation to an identity matrix.

As illustrated in ~\tabref{tab:quant_global}, our method achieves the best performance on all metrics.
Compared to WHAM, we can better handle errors in relative camera rotation estimation. On the EMDB dataset with dynamic camera inputs, using DPVO instead of Gyro results in only a 1.6mm/0.1\% drop in the W-MPJPE$_{100}$/RTE metrics, while WHAM experiences a drop of 19.5mm/1.9\%.
Compared to optimization-based algorithms like GLAMR and SLAHMR, our method also achieves better smoothness metrics. Although these methods incorporate a smoothness loss, they may struggle due to the high difficulty of the actions in the dataset.
Compared to regression methods like TRACE, our algorithm generalizes better to new datasets and achieves superior results.
An important baseline is HMR2.0+DPVO. We found that, although HMR2.0 performs well in camera-space~\tabref{tab:quant_incam}, it performs poorly in global motion recovery.
Particularly on the RICH dataset, the camera transformation is identity, indicating that camera-space estimation of human pose struggles to recover correct and consistent translation and scale.
Additionally, such methods cannot achieve gravity-aligned results.
In contrast, our algorithm naturally provides gravity-aligned results.

\begin{figure*}
    \centering
    \includegraphics[width=0.8\linewidth]{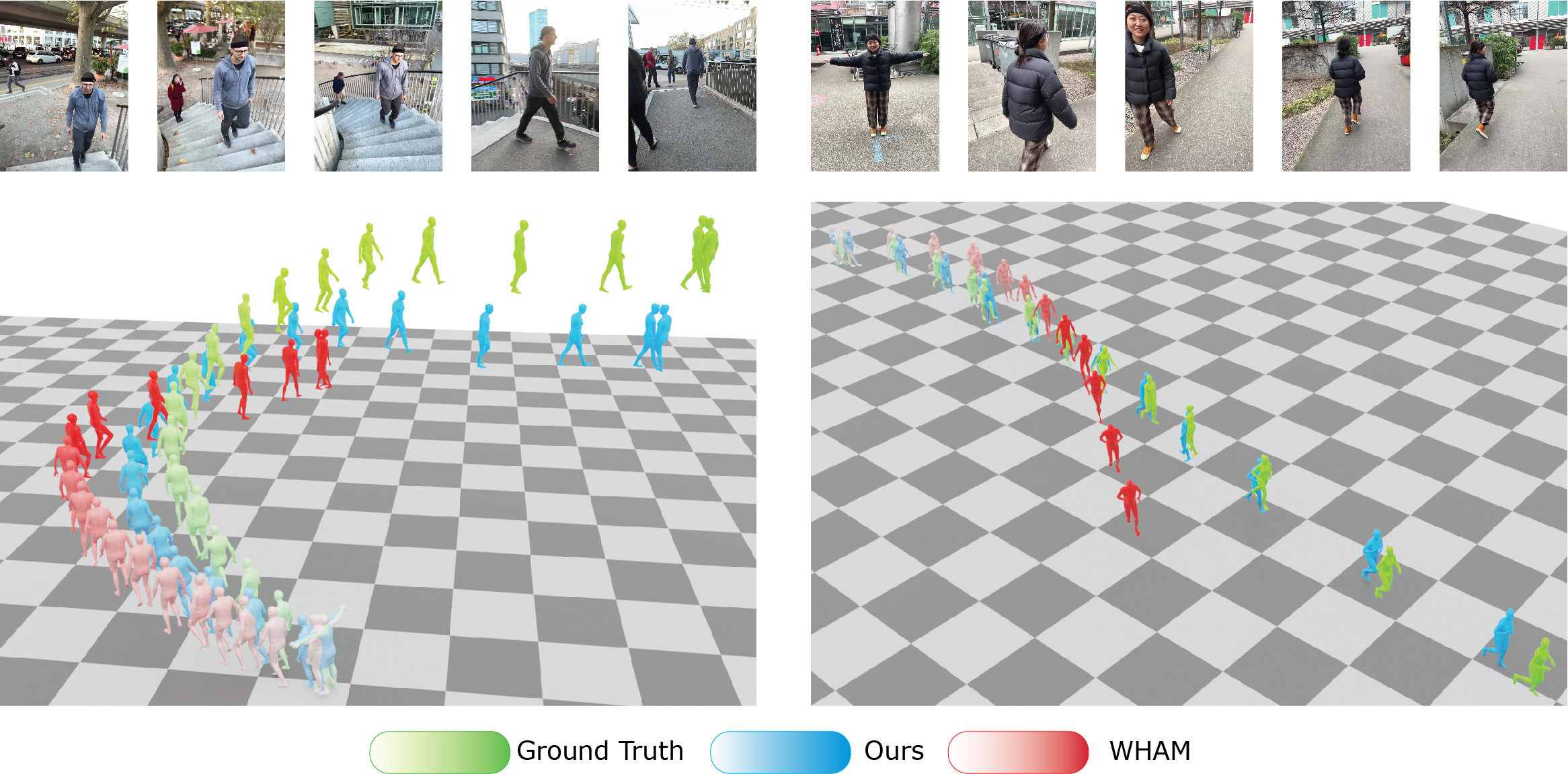}    
    \caption{\textbf{Qualitative results of global motion.} Our approach produces more accurate global motion than WHAM~\cite{wham}.}
    \label{fig:quality}
\end{figure*}

\begin{figure*}
    \centering
    \includegraphics[width=\linewidth]{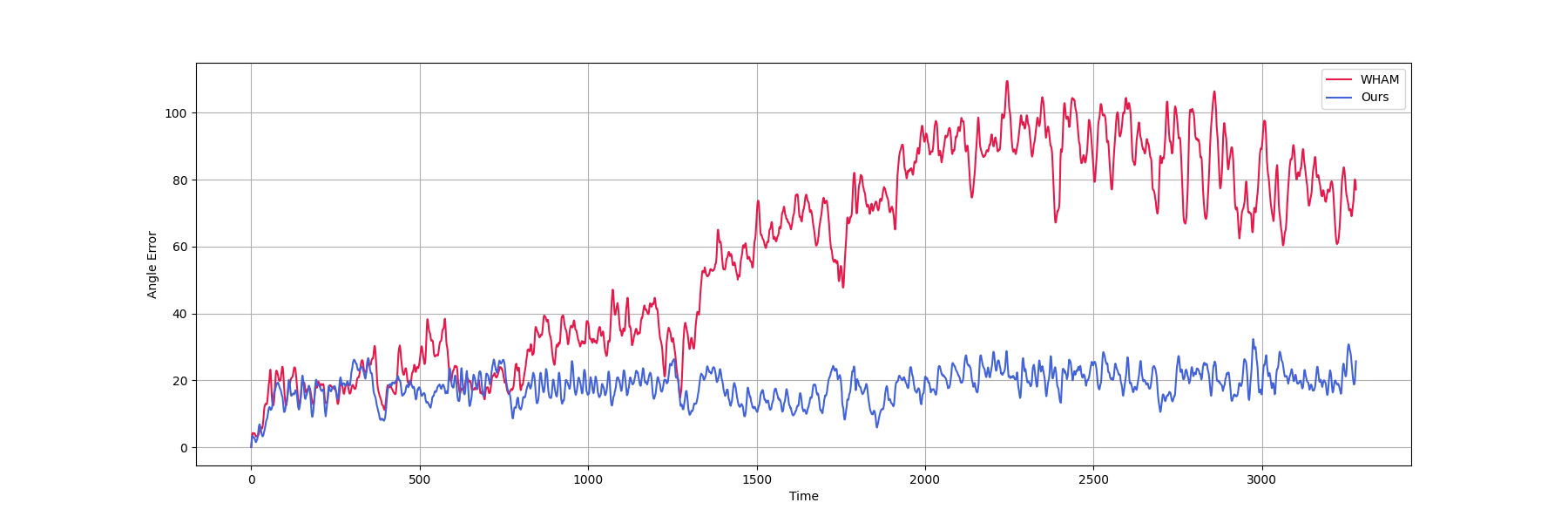}    
    \caption{\textbf{Global orientation error along time.} WHAM~\cite{wham} tends to accumulate more global orientation error as the sequence length increases, while our approach maintains a much lower error rate.}
    \label{fig:orierror}
\end{figure*}

As shown in ~\figref{fig:quality}, our method can recover more plausible global motion than WHAM.
To validate the effectiveness of our method, we show the global orientation angle error curve in~\figref{fig:orierror}. 
It can be observed that our method maintains a much lower error than WHAM, especially in the long-term prediction.

\subsection{Comparison on Camera Space Motion Recovery}
\begin{figure*}
    \centering
    \includegraphics[width=0.65\linewidth]{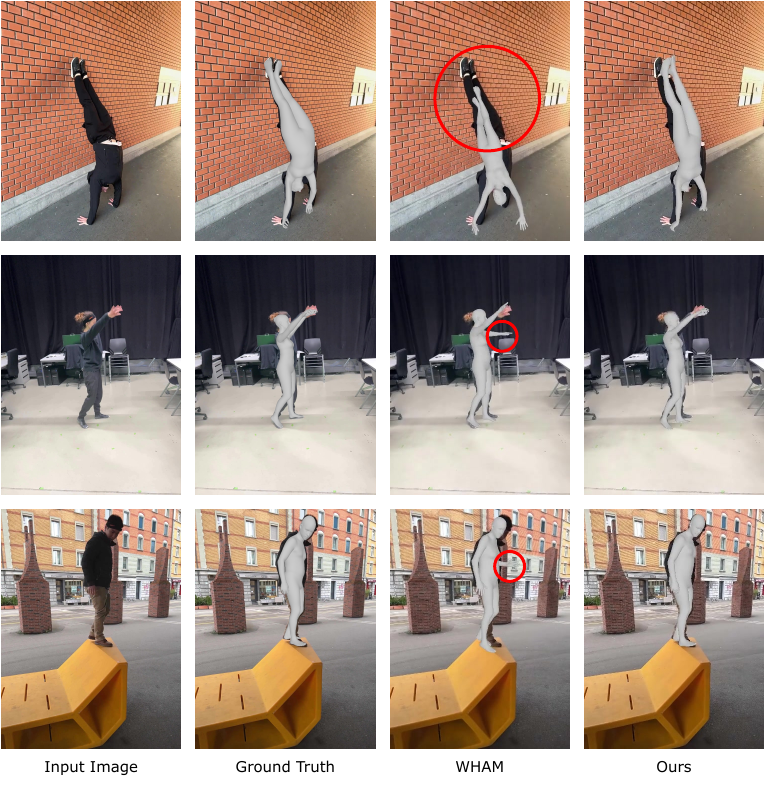}
    \caption{\textbf{Qualitative results of motion in camera coordinates.} WHAM~\cite{wham} could produce wrong results and fail to capture difficult motion (highlighted with red circles) while our approach could predict more plausible results.}
    \label{fig:qualitycam}
\end{figure*}

We compare our method with state-of-the-art motion recovery methods that predict camera-space results.
The results are shown in \tabref{tab:quant_incam}, where our method achieves the best performance on most of the metrics with a clear margin, demonstrating the effectiveness of our method in camera-space motion recovery.
We attribute this to the multitask learning strategy that enables our model to use global motion information to improve the camera-space motion estimation, especially the shape and smoothness of the motion.
Our PA-MPJPE performance is slightly behind WHAM by 0.3 mm on the 3DPW dataset. 
This may be due to the fact that we do not directly predict the SMPL parameters, but rather the SMPLX parameters, which might introduce some errors.
Nevertheless, the numbers are still competitive.
~\figref{fig:qualitycam} demonstrates that our approach estimates human motion in the camera space more accurately than WHAM.

\subsection{Understanding \methodname}
\begin{table}
    \centering
    \caption{\textbf{Ablation studies}. 
    We compare our method with seven variants on the RICH~\cite{rich} dataset (Refer to Sec.~4.4 for details).
    $^{*}$ denotes the variant that employs the sliding window.}
    \vspace{-2mm}
    \resizebox{\linewidth}{!}{
        \begin{tabular}{l|ccccccc}
            \toprule
            Variant                        & {PA-MPJPE} & {MPJPE}   & {WA-MPJPE} & {W-MPJPE}  & {RTE}    & {Jitter} & {Foot-Sliding} \\
            \midrule
            (1) w/o $GV$                   & 40.0       & 67.0     & 162.6	&278.9	&5.9	&9.7	&7.5           \\
            (2) w/o $\Gamma_{GV}$          & 41.4       & 70.5     & 101.2	&177.5	&4.5	&14.9	&3.0           \\
            (3) w/o Transformer            & 43.3       & 73.9     & 85.8	&138.9	&2.7	&\bf{7.6}	&3.3           \\
            (4) w/o $\text{Transformer}^*$ & 43.0       & 72.9     & 84.2	&142.0	&2.7	&10.6	&3.2           \\
            (5) w/o RoPE                   & 87.5       & 172.9    & 191.5	&304.4	&6.3	&22.8	&11.5           \\
            (6) w/o $\text{RoPE}^*$        & 40.1       & 67.9     & 80.7	&133.2	&2.4	&17.5	&3.3           \\
            (7) w/o PostProcessing         & 39.5       & 66.0     & 89.3	&145.2	&3.0	&14.5	&6.8           \\
            \midrule
            Full Model                     & \bf{39.5}  & \bf{66.0} & \bf{78.8}  & \bf{126.3} & \bf{2.4} & 12.8     & \bf{3.0}       \\
            \bottomrule
        \end{tabular}
        \label{tab:ablation_rich}
    }
\end{table}

\begin{figure*}
    \centering
    \includegraphics[width=\linewidth]{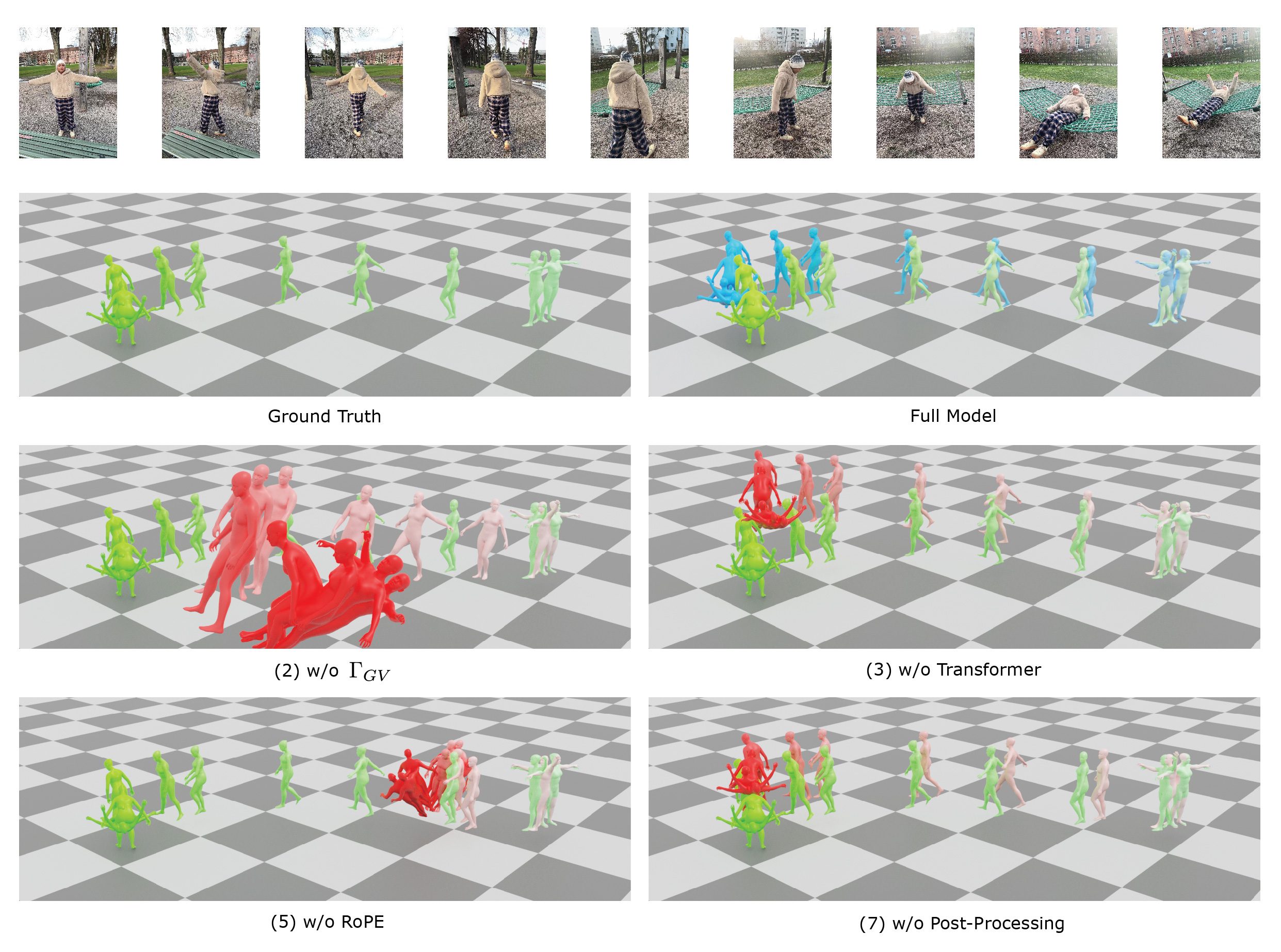}    
    \caption{\textbf{Qualitative results of ablations.} Each component of our method contributes to the final results.}
    \label{fig:ablation}
\end{figure*}

\paragraph{Ablation Studies.}
To understand the impact of each component in our method, we evaluate seven variants of \methodname using the same training and evaluation protocol on the RICH dataset.
The results are shown in \tabref{tab:ablation_rich}:
(1) \emph{w/o $GV$}: 
when predicting human motion solely in the camera coordinate system, the metrics drop slightly.
This suggests that gravity alignment improves camera-space human motion estimation accuracy.
For this variant, we can further recover a non-gravity-aligned global motion, which performs poorly in global metrics.
(2) \emph{w/o $\Gamma_{GV}$}: 
when predicting the relative global orientation from frame to frame,
the world-coordinate metrics drop substantially, indicating that the model suffers from error accumulation in this configuration.
(3) \emph{w/o Transformer}: adopting a convolutional architecture yields poor performance, highlighting that our transformer architecture is more effective.
(4) \emph{w/o $\text{Transformer}^*$}: when applying a convolutional architecture with a sliding window inference strategy, the performance remains similarly poor, further validating the superiority of our transformer approach.
(5) \emph{w/o RoPE}: substituting RoPE with absolute positional encoding leads to very poor results.
This is primarily because absolute positional embedding struggles to generalize well in long sequences.
(6) \emph{w/o $\text{RoPE}^*$}: even when using absolute positional embedding with a sliding window inference strategy, the results are still worse than our approach, confirming the inadequacy of this embedding strategy.
(7) \emph{w/o Post-Processing}: Omitting the postprocessing step causes a significant increase in global metrics, demonstrating that our postprocessing strategy substantially enhances global accuracy.
~\figref{fig:ablation} demonstrates that each component of our approach contributes to the overall performance.
We find similar conclusion on the EMDB dataset, which is presented in the supplementary material.

\begin{table}
    \centering
    \caption{\textbf{Dataset and test-time-augmentation ablation on EMDB}. 
    B denotes BEDLAM~\cite{bedlam} synthetic dataset.}
    \vspace{-2mm}
    \resizebox{0.9\linewidth}{!}{
        \begin{tabular}{l|ccc|ccc}
            \toprule
            Method                      & {PA-MPJPE} & {MPJPE}  & {Accel}  & {WA-MPJPE}& {W-MPJPE} &{RTE}\\
            \midrule
            \text{WHAM +B}&49.4&78.2&6.0&134.2&338.1&3.8  \\ 
            \text{WHAM +B+FlipEval}&47.9&76.9&5.4&132.5&337.7&3.8\\ 
            \text{\methodname+B}&44.2&74.0&4.0&110.6&274.9&1.9\\ 
            \text{\methodname+B+FlipEval}&42.7&72.6&3.6&109.1&272.9&1.9\\
            \bottomrule
        \end{tabular}
        \label{tab:ablation_dataset}
    }
\end{table}
In ~\tabref{tab:ablation_dataset}, we provide a comparison with the most relevant baseline method, WHAM. 
When trained on the BEDLAM dataset, with or without using FlipEval as a test-time augmentation, \methodname shows a significant performance improvement over WHAM. Additionally, we observe that FlipEval offers greater improvements in camera-space metrics compared to global-space metrics.

\paragraph{Running Time.}
We test the running time with an example video of 1430 frames (approximately 45 seconds).
The preprocessing, which includes YOLOv8 detection, ViTPose, Vit feature extraction, and DPVO, takes a total of 46.0 seconds $(4.9 + 20.0 + 10.1 + 11.0)$. The rest of the \methodname takes 0.28 seconds.
WHAM adopts the same preprocessing procedures, and it requires 2.0 seconds for the core network.
The optimization-based method SLAHMR takes more than 6 hours to process. All models are tested with an RTX 4090 GPU.
The improved efficiency enables scalable processing of human motion videos, aiding in the creation of foundational datasets.

\section{Conclusions}
We introduce \methodname, a novel approach for regressing world-grounded human motion from monocular videos.
\methodname defines a Gravity-View (GV) coordinate system to leverage gravity priors and constraints, avoiding error accumulation along the gravity axis.
By incorporating a relative transformer with RoPE, \methodname handles sequences of arbitrary length during inference, without the need for sliding-window. 
Extensive experiments demonstrate that \methodname outperforms existing methods across various benchmarks, achieving state-of-the-art accuracy and motion plausibility in both camera-space and world-grounded metrics.

\section*{Acknowledgements}
The authors would like to acknowledge support from NSFC (No. 62172364), Information Technology Center and State Key Lab of CAD\&CG, Zhejiang University.

\newpage

\bibliographystyle{ACM-Reference-Format}
\bibliography{ref}


\begin{thebibliography}{51}


\ifx \showCODEN    \undefined \def \showCODEN     #1{\unskip}     \fi
\ifx \showDOI      \undefined \def \showDOI       #1{#1}\fi
\ifx \showISBNx    \undefined \def \showISBNx     #1{\unskip}     \fi
\ifx \showISBNxiii \undefined \def \showISBNxiii  #1{\unskip}     \fi
\ifx \showISSN     \undefined \def \showISSN      #1{\unskip}     \fi
\ifx \showLCCN     \undefined \def \showLCCN      #1{\unskip}     \fi
\ifx \shownote     \undefined \def \shownote      #1{#1}          \fi
\ifx \showarticletitle \undefined \def \showarticletitle #1{#1}   \fi
\ifx \showURL      \undefined \def \showURL       {\relax}        \fi
\providecommand\bibfield[2]{#2}
\providecommand\bibinfo[2]{#2}
\providecommand\natexlab[1]{#1}
\providecommand\showeprint[2][]{arXiv:#2}

\bibitem[\protect\citeauthoryear{Aristidou and Lasenby}{Aristidou and
  Lasenby}{2011}]%
        {ccdik}
\bibfield{author}{\bibinfo{person}{Andreas Aristidou} {and}
  \bibinfo{person}{Joan Lasenby}.} \bibinfo{year}{2011}\natexlab{}.
\newblock \showarticletitle{FABRIK: A fast, iterative solver for the Inverse
  Kinematics problem}.
\newblock \bibinfo{journal}{\emph{Graphical Models}} \bibinfo{volume}{73},
  \bibinfo{number}{5} (\bibinfo{year}{2011}), \bibinfo{pages}{243--260}.
\newblock


\bibitem[\protect\citeauthoryear{Black, Patel, Tesch, and Yang}{Black
  et~al\mbox{.}}{2023}]%
        {bedlam}
\bibfield{author}{\bibinfo{person}{Michael~J. Black}, \bibinfo{person}{Priyanka
  Patel}, \bibinfo{person}{Joachim Tesch}, {and} \bibinfo{person}{Jinlong
  Yang}.} \bibinfo{year}{2023}\natexlab{}.
\newblock \showarticletitle{{BEDLAM}: A Synthetic Dataset of Bodies Exhibiting
  Detailed Lifelike Animated Motion}. In \bibinfo{booktitle}{\emph{Proceedings
  IEEE/CVF Conf.~on Computer Vision and Pattern Recognition (CVPR)}}.
  \bibinfo{pages}{8726--8737}.
\newblock


\bibitem[\protect\citeauthoryear{Bogo, Kanazawa, Lassner, Gehler, Romero, and
  Black}{Bogo et~al\mbox{.}}{2016}]%
        {smplify}
\bibfield{author}{\bibinfo{person}{Federica Bogo}, \bibinfo{person}{Angjoo
  Kanazawa}, \bibinfo{person}{Christoph Lassner}, \bibinfo{person}{Peter
  Gehler}, \bibinfo{person}{Javier Romero}, {and} \bibinfo{person}{Michael~J.
  Black}.} \bibinfo{year}{2016}\natexlab{}.
\newblock \showarticletitle{Keep it {SMPL}: Automatic Estimation of {3D} Human
  Pose and Shape from a Single Image}. In \bibinfo{booktitle}{\emph{Computer
  Vision -- ECCV 2016}} \emph{(\bibinfo{series}{Lecture Notes in Computer
  Science})}. \bibinfo{publisher}{Springer International Publishing},
  \bibinfo{pages}{561--578}.
\newblock


\bibitem[\protect\citeauthoryear{Choi, Moon, Chang, and Lee}{Choi
  et~al\mbox{.}}{2021}]%
        {tcmr}
\bibfield{author}{\bibinfo{person}{Hongsuk Choi}, \bibinfo{person}{Gyeongsik
  Moon}, \bibinfo{person}{Ju~Yong Chang}, {and} \bibinfo{person}{Kyoung~Mu
  Lee}.} \bibinfo{year}{2021}\natexlab{}.
\newblock \showarticletitle{Beyond Static Features for Temporally Consistent 3D
  Human Pose and Shape from a Video}. In \bibinfo{booktitle}{\emph{Conference
  on Computer Vision and Pattern Recognition (CVPR)}}.
\newblock


\bibitem[\protect\citeauthoryear{Goel, Pavlakos, Rajasegaran, Kanazawa, and
  Malik}{Goel et~al\mbox{.}}{2023}]%
        {hmr2}
\bibfield{author}{\bibinfo{person}{Shubham Goel}, \bibinfo{person}{Georgios
  Pavlakos}, \bibinfo{person}{Jathushan Rajasegaran}, \bibinfo{person}{Angjoo
  Kanazawa}, {and} \bibinfo{person}{Jitendra Malik}.}
  \bibinfo{year}{2023}\natexlab{}.
\newblock \showarticletitle{Humans in 4D: Reconstructing and tracking humans
  with transformers}. In \bibinfo{booktitle}{\emph{Proceedings of the IEEE/CVF
  International Conference on Computer Vision}}. \bibinfo{pages}{14783--14794}.
\newblock


\bibitem[\protect\citeauthoryear{Guo, Zou, Zuo, Wang, Ji, Li, and Cheng}{Guo
  et~al\mbox{.}}{2022}]%
        {t2m}
\bibfield{author}{\bibinfo{person}{Chuan Guo}, \bibinfo{person}{Shihao Zou},
  \bibinfo{person}{Xinxin Zuo}, \bibinfo{person}{Sen Wang},
  \bibinfo{person}{Wei Ji}, \bibinfo{person}{Xingyu Li}, {and}
  \bibinfo{person}{Li Cheng}.} \bibinfo{year}{2022}\natexlab{}.
\newblock \showarticletitle{Generating diverse and natural 3d human motions
  from text}. In \bibinfo{booktitle}{\emph{Proceedings of the IEEE/CVF
  Conference on Computer Vision and Pattern Recognition}}.
  \bibinfo{pages}{5152--5161}.
\newblock


\bibitem[\protect\citeauthoryear{He, Luo, Xiao, Zhang, Kitani, Liu, and Shi}{He
  et~al\mbox{.}}{2024}]%
        {human2humanoid}
\bibfield{author}{\bibinfo{person}{Tairan He}, \bibinfo{person}{Zhengyi Luo},
  \bibinfo{person}{Wenli Xiao}, \bibinfo{person}{Chong Zhang},
  \bibinfo{person}{Kris Kitani}, \bibinfo{person}{Changliu Liu}, {and}
  \bibinfo{person}{Guanya Shi}.} \bibinfo{year}{2024}\natexlab{}.
\newblock \showarticletitle{Learning Human-to-Humanoid Real-Time Whole-Body
  Teleoperation}. In \bibinfo{booktitle}{\emph{arXiv}}.
\newblock


\bibitem[\protect\citeauthoryear{Huang, Yi, H{\"o}schle, Safroshkin, Alexiadis,
  Polikovsky, Scharstein, and Black}{Huang et~al\mbox{.}}{2022}]%
        {rich}
\bibfield{author}{\bibinfo{person}{Chun-Hao~P. Huang}, \bibinfo{person}{Hongwei
  Yi}, \bibinfo{person}{Markus H{\"o}schle}, \bibinfo{person}{Matvey
  Safroshkin}, \bibinfo{person}{Tsvetelina Alexiadis}, \bibinfo{person}{Senya
  Polikovsky}, \bibinfo{person}{Daniel Scharstein}, {and}
  \bibinfo{person}{Michael~J. Black}.} \bibinfo{year}{2022}\natexlab{}.
\newblock \showarticletitle{Capturing and Inferring Dense Full-Body Human-Scene
  Contact}. In \bibinfo{booktitle}{\emph{Proceedings IEEE/CVF Conf.~on Computer
  Vision and Pattern Recognition (CVPR)}}. \bibinfo{pages}{13274--13285}.
\newblock


\bibitem[\protect\citeauthoryear{Ionescu, Papava, Olaru, and
  Sminchisescu}{Ionescu et~al\mbox{.}}{2014}]%
        {h36m}
\bibfield{author}{\bibinfo{person}{Catalin Ionescu}, \bibinfo{person}{Dragos
  Papava}, \bibinfo{person}{Vlad Olaru}, {and} \bibinfo{person}{Cristian
  Sminchisescu}.} \bibinfo{year}{2014}\natexlab{}.
\newblock \showarticletitle{Human3.6M: Large Scale Datasets and Predictive
  Methods for 3D Human Sensing in Natural Environments}.
\newblock \bibinfo{journal}{\emph{IEEE Transactions on Pattern Analysis and
  Machine Intelligence}} \bibinfo{volume}{36}, \bibinfo{number}{7}
  (\bibinfo{date}{jul} \bibinfo{year}{2014}), \bibinfo{pages}{1325--1339}.
\newblock


\bibitem[\protect\citeauthoryear{Jocher, Chaurasia, and Qiu}{Jocher
  et~al\mbox{.}}{2023}]%
        {yolov8}
\bibfield{author}{\bibinfo{person}{Glenn Jocher}, \bibinfo{person}{Ayush
  Chaurasia}, {and} \bibinfo{person}{Jing Qiu}.}
  \bibinfo{year}{2023}\natexlab{}.
\newblock \bibinfo{booktitle}{\emph{Ultralytics YOLOv8}}.
\newblock
\urldef\tempurl%
\url{https://github.com/ultralytics/ultralytics}
\showURL{%
\tempurl}


\bibitem[\protect\citeauthoryear{Kanazawa, Black, Jacobs, and Malik}{Kanazawa
  et~al\mbox{.}}{2018}]%
        {hmr}
\bibfield{author}{\bibinfo{person}{Angjoo Kanazawa},
  \bibinfo{person}{Michael~J. Black}, \bibinfo{person}{David~W. Jacobs}, {and}
  \bibinfo{person}{Jitendra Malik}.} \bibinfo{year}{2018}\natexlab{}.
\newblock \showarticletitle{End-to-end Recovery of Human Shape and Pose}. In
  \bibinfo{booktitle}{\emph{Computer Vision and Pattern Regognition (CVPR)}}.
\newblock


\bibitem[\protect\citeauthoryear{Kanazawa, Zhang, Felsen, and Malik}{Kanazawa
  et~al\mbox{.}}{2019}]%
        {hmmr}
\bibfield{author}{\bibinfo{person}{Angjoo Kanazawa}, \bibinfo{person}{Jason~Y
  Zhang}, \bibinfo{person}{Panna Felsen}, {and} \bibinfo{person}{Jitendra
  Malik}.} \bibinfo{year}{2019}\natexlab{}.
\newblock \showarticletitle{Learning 3d human dynamics from video}. In
  \bibinfo{booktitle}{\emph{Proceedings of the IEEE/CVF conference on computer
  vision and pattern recognition}}. \bibinfo{pages}{5614--5623}.
\newblock


\bibitem[\protect\citeauthoryear{Kaufmann, Song, Guo, Shen, Jiang, Tang,
  Z{\'a}rate, and Hilliges}{Kaufmann et~al\mbox{.}}{2023}]%
        {emdb}
\bibfield{author}{\bibinfo{person}{Manuel Kaufmann}, \bibinfo{person}{Jie
  Song}, \bibinfo{person}{Chen Guo}, \bibinfo{person}{Kaiyue Shen},
  \bibinfo{person}{Tianjian Jiang}, \bibinfo{person}{Chengcheng Tang},
  \bibinfo{person}{Juan~Jos{\'e} Z{\'a}rate}, {and} \bibinfo{person}{Otmar
  Hilliges}.} \bibinfo{year}{2023}\natexlab{}.
\newblock \showarticletitle{{EMDB}: The {E}lectromagnetic {D}atabase of
  {G}lobal 3{D} {H}uman {P}ose and {S}hape in the {W}ild}. In
  \bibinfo{booktitle}{\emph{International Conference on Computer Vision
  (ICCV)}}.
\newblock


\bibitem[\protect\citeauthoryear{Kocabas, Athanasiou, and Black}{Kocabas
  et~al\mbox{.}}{2020}]%
        {vibe}
\bibfield{author}{\bibinfo{person}{Muhammed Kocabas}, \bibinfo{person}{Nikos
  Athanasiou}, {and} \bibinfo{person}{Michael~J. Black}.}
  \bibinfo{year}{2020}\natexlab{}.
\newblock \showarticletitle{{VIBE}: Video Inference for Human Body Pose and
  Shape Estimation}. In \bibinfo{booktitle}{\emph{Proceedings IEEE Conf. on
  Computer Vision and Pattern Recognition (CVPR)}}. \bibinfo{publisher}{IEEE},
  \bibinfo{pages}{5252--5262}.
\newblock
\urldef\tempurl%
\url{https://doi.org/10.1109/CVPR42600.2020.00530}
\showDOI{\tempurl}


\bibitem[\protect\citeauthoryear{Kocabas, Huang, Hilliges, and Black}{Kocabas
  et~al\mbox{.}}{2021a}]%
        {pare}
\bibfield{author}{\bibinfo{person}{Muhammed Kocabas},
  \bibinfo{person}{Chun-Hao~P. Huang}, \bibinfo{person}{Otmar Hilliges}, {and}
  \bibinfo{person}{Michael~J. Black}.} \bibinfo{year}{2021}\natexlab{a}.
\newblock \showarticletitle{{PARE}: Part Attention Regressor for {3D} Human
  Body Estimation}. In \bibinfo{booktitle}{\emph{Proc. International Conference
  on Computer Vision (ICCV)}}. \bibinfo{pages}{11127--11137}.
\newblock


\bibitem[\protect\citeauthoryear{Kocabas, Huang, Tesch, M\"uller, Hilliges, and
  Black}{Kocabas et~al\mbox{.}}{2021b}]%
        {spec}
\bibfield{author}{\bibinfo{person}{Muhammed Kocabas},
  \bibinfo{person}{Chun-Hao~P. Huang}, \bibinfo{person}{Joachim Tesch},
  \bibinfo{person}{Lea M\"uller}, \bibinfo{person}{Otmar Hilliges}, {and}
  \bibinfo{person}{Michael~J. Black}.} \bibinfo{year}{2021}\natexlab{b}.
\newblock \showarticletitle{{SPEC}: Seeing People in the Wild with an Estimated
  Camera}. In \bibinfo{booktitle}{\emph{Proc. International Conference on
  Computer Vision (ICCV)}}. \bibinfo{publisher}{IEEE},
  \bibinfo{address}{Piscataway, NJ}, \bibinfo{pages}{11015--11025}.
\newblock
\urldef\tempurl%
\url{https://doi.org/10.1109/ICCV48922.2021.01085}
\showDOI{\tempurl}


\bibitem[\protect\citeauthoryear{Kocabas, Yuan, Molchanov, Guo, Black,
  Hilliges, Kautz, and Iqbal}{Kocabas et~al\mbox{.}}{2024}]%
        {pace}
\bibfield{author}{\bibinfo{person}{Muhammed Kocabas}, \bibinfo{person}{Ye
  Yuan}, \bibinfo{person}{Pavlo Molchanov}, \bibinfo{person}{Yunrong Guo},
  \bibinfo{person}{Michael~J. Black}, \bibinfo{person}{Otmar Hilliges},
  \bibinfo{person}{Jan Kautz}, {and} \bibinfo{person}{Umar Iqbal}.}
  \bibinfo{year}{2024}\natexlab{}.
\newblock \showarticletitle{PACE: Human and Motion Estimation from in-the-wild
  Videos}. In \bibinfo{booktitle}{\emph{3DV}}.
\newblock


\bibitem[\protect\citeauthoryear{Kolotouros, Pavlakos, Black, and
  Daniilidis}{Kolotouros et~al\mbox{.}}{2019}]%
        {spin}
\bibfield{author}{\bibinfo{person}{Nikos Kolotouros}, \bibinfo{person}{Georgios
  Pavlakos}, \bibinfo{person}{Michael~J Black}, {and} \bibinfo{person}{Kostas
  Daniilidis}.} \bibinfo{year}{2019}\natexlab{}.
\newblock \showarticletitle{Learning to Reconstruct 3D Human Pose and Shape via
  Model-fitting in the Loop}. In \bibinfo{booktitle}{\emph{ICCV}}.
\newblock


\bibitem[\protect\citeauthoryear{Li, Bian, Liu, Tang, Wang, and Lu}{Li
  et~al\mbox{.}}{2023}]%
        {niki}
\bibfield{author}{\bibinfo{person}{Jiefeng Li}, \bibinfo{person}{Siyuan Bian},
  \bibinfo{person}{Qi Liu}, \bibinfo{person}{Jiasheng Tang},
  \bibinfo{person}{Fan Wang}, {and} \bibinfo{person}{Cewu Lu}.}
  \bibinfo{year}{2023}\natexlab{}.
\newblock \showarticletitle{{NIKI}: Neural Inverse Kinematics with Invertible
  Neural Networks for 3D Human Pose and Shape Estimation}. In
  \bibinfo{booktitle}{\emph{Proceedings of the IEEE/CVF Conference on Computer
  Vision and Pattern Recognition (CVPR)}}.
\newblock


\bibitem[\protect\citeauthoryear{Li, Bian, Xu, Liu, Yu, and Lu}{Li
  et~al\mbox{.}}{2022a}]%
        {dandd}
\bibfield{author}{\bibinfo{person}{Jiefeng Li}, \bibinfo{person}{Siyuan Bian},
  \bibinfo{person}{Chao Xu}, \bibinfo{person}{Gang Liu}, \bibinfo{person}{Gang
  Yu}, {and} \bibinfo{person}{Cewu Lu}.} \bibinfo{year}{2022}\natexlab{a}.
\newblock \showarticletitle{D \&d: Learning human dynamics from dynamic
  camera}. In \bibinfo{booktitle}{\emph{European Conference on Computer
  Vision}}. Springer, \bibinfo{pages}{479--496}.
\newblock


\bibitem[\protect\citeauthoryear{Li, Xu, Chen, Bian, Yang, and Lu}{Li
  et~al\mbox{.}}{2021}]%
        {hybrik}
\bibfield{author}{\bibinfo{person}{Jiefeng Li}, \bibinfo{person}{Chao Xu},
  \bibinfo{person}{Zhicun Chen}, \bibinfo{person}{Siyuan Bian},
  \bibinfo{person}{Lixin Yang}, {and} \bibinfo{person}{Cewu Lu}.}
  \bibinfo{year}{2021}\natexlab{}.
\newblock \showarticletitle{Hybrik: A hybrid analytical-neural inverse
  kinematics solution for 3d human pose and shape estimation}. In
  \bibinfo{booktitle}{\emph{Proceedings of the IEEE/CVF Conference on Computer
  Vision and Pattern Recognition}}. \bibinfo{pages}{3383--3393}.
\newblock


\bibitem[\protect\citeauthoryear{Li, Liu, Zhang, Xu, and Yan}{Li
  et~al\mbox{.}}{2022b}]%
        {cliff}
\bibfield{author}{\bibinfo{person}{Zhihao Li}, \bibinfo{person}{Jianzhuang
  Liu}, \bibinfo{person}{Zhensong Zhang}, \bibinfo{person}{Songcen Xu}, {and}
  \bibinfo{person}{Youliang Yan}.} \bibinfo{year}{2022}\natexlab{b}.
\newblock \showarticletitle{CLIFF: Carrying Location Information in Full Frames
  into Human Pose and Shape Estimation}. In \bibinfo{booktitle}{\emph{ECCV}}.
\newblock


\bibitem[\protect\citeauthoryear{Loper, Mahmood, Romero, Pons-Moll, and
  Black}{Loper et~al\mbox{.}}{2023}]%
        {smpl}
\bibfield{author}{\bibinfo{person}{Matthew Loper}, \bibinfo{person}{Naureen
  Mahmood}, \bibinfo{person}{Javier Romero}, \bibinfo{person}{Gerard
  Pons-Moll}, {and} \bibinfo{person}{Michael~J Black}.}
  \bibinfo{year}{2023}\natexlab{}.
\newblock \showarticletitle{SMPL: A skinned multi-person linear model}.
\newblock In \bibinfo{booktitle}{\emph{Seminal Graphics Papers: Pushing the
  Boundaries, Volume 2}}. \bibinfo{pages}{851--866}.
\newblock


\bibitem[\protect\citeauthoryear{Luo, Golestaneh, and Kitani}{Luo
  et~al\mbox{.}}{2020}]%
        {meva}
\bibfield{author}{\bibinfo{person}{Zhengyi Luo}, \bibinfo{person}{S.~Alireza
  Golestaneh}, {and} \bibinfo{person}{Kris~M. Kitani}.}
  \bibinfo{year}{2020}\natexlab{}.
\newblock \showarticletitle{3D Human Motion Estimation via Motion Compression
  and Refinement}. In \bibinfo{booktitle}{\emph{Proceedings of the Asian
  Conference on Computer Vision (ACCV)}}.
\newblock


\bibitem[\protect\citeauthoryear{Mahmood, Ghorbani, Troje, Pons-Moll, and
  Black}{Mahmood et~al\mbox{.}}{2019}]%
        {amass}
\bibfield{author}{\bibinfo{person}{Naureen Mahmood}, \bibinfo{person}{Nima
  Ghorbani}, \bibinfo{person}{Nikolaus~F. Troje}, \bibinfo{person}{Gerard
  Pons-Moll}, {and} \bibinfo{person}{Michael~J. Black}.}
  \bibinfo{year}{2019}\natexlab{}.
\newblock \showarticletitle{{AMASS}: Archive of Motion Capture as Surface
  Shapes}. In \bibinfo{booktitle}{\emph{International Conference on Computer
  Vision}}. \bibinfo{pages}{5442--5451}.
\newblock


\bibitem[\protect\citeauthoryear{Pavlakos, Choutas, Ghorbani, Bolkart, Osman,
  Tzionas, and Black}{Pavlakos et~al\mbox{.}}{2019}]%
        {smplx}
\bibfield{author}{\bibinfo{person}{Georgios Pavlakos},
  \bibinfo{person}{Vasileios Choutas}, \bibinfo{person}{Nima Ghorbani},
  \bibinfo{person}{Timo Bolkart}, \bibinfo{person}{Ahmed A.~A. Osman},
  \bibinfo{person}{Dimitrios Tzionas}, {and} \bibinfo{person}{Michael~J.
  Black}.} \bibinfo{year}{2019}\natexlab{}.
\newblock \showarticletitle{Expressive Body Capture: {3D} Hands, Face, and Body
  from a Single Image}. In \bibinfo{booktitle}{\emph{Proceedings IEEE Conf. on
  Computer Vision and Pattern Recognition (CVPR)}}.
  \bibinfo{pages}{10975--10985}.
\newblock


\bibitem[\protect\citeauthoryear{Press, Smith, and Lewis}{Press
  et~al\mbox{.}}{2022}]%
        {alibi}
\bibfield{author}{\bibinfo{person}{Ofir Press}, \bibinfo{person}{Noah Smith},
  {and} \bibinfo{person}{Mike Lewis}.} \bibinfo{year}{2022}\natexlab{}.
\newblock \showarticletitle{Train Short, Test Long: Attention with Linear
  Biases Enables Input Length Extrapolation}. In
  \bibinfo{booktitle}{\emph{International Conference on Learning
  Representations}}.
\newblock
\urldef\tempurl%
\url{https://openreview.net/forum?id=R8sQPpGCv0}
\showURL{%
\tempurl}


\bibitem[\protect\citeauthoryear{Rempe, Birdal, Hertzmann, Yang, Sridhar, and
  Guibas}{Rempe et~al\mbox{.}}{2021}]%
        {humor}
\bibfield{author}{\bibinfo{person}{Davis Rempe}, \bibinfo{person}{Tolga
  Birdal}, \bibinfo{person}{Aaron Hertzmann}, \bibinfo{person}{Jimei Yang},
  \bibinfo{person}{Srinath Sridhar}, {and} \bibinfo{person}{Leonidas~J.
  Guibas}.} \bibinfo{year}{2021}\natexlab{}.
\newblock \showarticletitle{HuMoR: 3D Human Motion Model for Robust Pose
  Estimation}. In \bibinfo{booktitle}{\emph{International Conference on
  Computer Vision (ICCV)}}.
\newblock


\bibitem[\protect\citeauthoryear{Shen, Yang, Wang, Ma, Zhou, and Yang}{Shen
  et~al\mbox{.}}{2023}]%
        {glot}
\bibfield{author}{\bibinfo{person}{Xiaolong Shen}, \bibinfo{person}{Zongxin
  Yang}, \bibinfo{person}{Xiaohan Wang}, \bibinfo{person}{Jianxin Ma},
  \bibinfo{person}{Chang Zhou}, {and} \bibinfo{person}{Yi Yang}.}
  \bibinfo{year}{2023}\natexlab{}.
\newblock \showarticletitle{Global-to-Local Modeling for Video-Based 3D Human
  Pose and Shape Estimation}. In \bibinfo{booktitle}{\emph{Proceedings of the
  IEEE/CVF Conference on Computer Vision and Pattern Recognition (CVPR)}}.
  \bibinfo{pages}{8887--8896}.
\newblock


\bibitem[\protect\citeauthoryear{Shi, Aberman, Aristidou, Komura, Lischinski,
  Cohen-Or, and Chen}{Shi et~al\mbox{.}}{2020}]%
        {shi2020motionet}
\bibfield{author}{\bibinfo{person}{Mingyi Shi}, \bibinfo{person}{Kfir Aberman},
  \bibinfo{person}{Andreas Aristidou}, \bibinfo{person}{Taku Komura},
  \bibinfo{person}{Dani Lischinski}, \bibinfo{person}{Daniel Cohen-Or}, {and}
  \bibinfo{person}{Baoquan Chen}.} \bibinfo{year}{2020}\natexlab{}.
\newblock \showarticletitle{Motionet: 3d human motion reconstruction from
  monocular video with skeleton consistency}.
\newblock \bibinfo{journal}{\emph{Acm transactions on graphics (tog)}}
  \bibinfo{volume}{40}, \bibinfo{number}{1} (\bibinfo{year}{2020}),
  \bibinfo{pages}{1--15}.
\newblock


\bibitem[\protect\citeauthoryear{Shin, Kim, Halilaj, and Black}{Shin
  et~al\mbox{.}}{2024}]%
        {wham}
\bibfield{author}{\bibinfo{person}{Soyong Shin}, \bibinfo{person}{Juyong Kim},
  \bibinfo{person}{Eni Halilaj}, {and} \bibinfo{person}{Michael~J Black}.}
  \bibinfo{year}{2024}\natexlab{}.
\newblock \showarticletitle{Wham: Reconstructing world-grounded humans with
  accurate 3d motion}. In \bibinfo{booktitle}{\emph{Proceedings of the IEEE/CVF
  Conference on Computer Vision and Pattern Recognition}}.
  \bibinfo{pages}{2070--2080}.
\newblock


\bibitem[\protect\citeauthoryear{Starke, Zhang, Komura, and Saito}{Starke
  et~al\mbox{.}}{2019}]%
        {nsm}
\bibfield{author}{\bibinfo{person}{Sebastian Starke}, \bibinfo{person}{He
  Zhang}, \bibinfo{person}{Taku Komura}, {and} \bibinfo{person}{Jun Saito}.}
  \bibinfo{year}{2019}\natexlab{}.
\newblock \showarticletitle{Neural state machine for character-scene
  interactions}.
\newblock \bibinfo{journal}{\emph{ACM Transactions on Graphics}}
  \bibinfo{volume}{38}, \bibinfo{number}{6} (\bibinfo{year}{2019}),
  \bibinfo{pages}{178}.
\newblock


\bibitem[\protect\citeauthoryear{Su, Ahmed, Lu, Pan, Bo, and Liu}{Su
  et~al\mbox{.}}{2024}]%
        {rope}
\bibfield{author}{\bibinfo{person}{Jianlin Su}, \bibinfo{person}{Murtadha
  Ahmed}, \bibinfo{person}{Yu Lu}, \bibinfo{person}{Shengfeng Pan},
  \bibinfo{person}{Wen Bo}, {and} \bibinfo{person}{Yunfeng Liu}.}
  \bibinfo{year}{2024}\natexlab{}.
\newblock \showarticletitle{Roformer: Enhanced transformer with rotary position
  embedding}.
\newblock \bibinfo{journal}{\emph{Neurocomputing}}  \bibinfo{volume}{568}
  (\bibinfo{year}{2024}), \bibinfo{pages}{127063}.
\newblock


\bibitem[\protect\citeauthoryear{Sun, Bao, Liu, Mei, and Black}{Sun
  et~al\mbox{.}}{2023}]%
        {trace}
\bibfield{author}{\bibinfo{person}{Yu Sun}, \bibinfo{person}{Qian Bao},
  \bibinfo{person}{Wu Liu}, \bibinfo{person}{Tao Mei}, {and}
  \bibinfo{person}{Michael~J. Black}.} \bibinfo{year}{2023}\natexlab{}.
\newblock \showarticletitle{{TRACE: 5D Temporal Regression of Avatars with
  Dynamic Cameras in 3D Environments}}. In \bibinfo{booktitle}{\emph{IEEE/CVF
  Conf.~on Computer Vision and Pattern Recognition (CVPR)}}.
\newblock


\bibitem[\protect\citeauthoryear{Sun, Ye, Liu, Gao, Fu, and Mei}{Sun
  et~al\mbox{.}}{2019}]%
        {dsd-satn}
\bibfield{author}{\bibinfo{person}{Yu Sun}, \bibinfo{person}{Yun Ye},
  \bibinfo{person}{Wu Liu}, \bibinfo{person}{Wenpeng Gao},
  \bibinfo{person}{YiLi Fu}, {and} \bibinfo{person}{Tao Mei}.}
  \bibinfo{year}{2019}\natexlab{}.
\newblock \showarticletitle{Human Mesh Recovery from Monocular Images via a
  Skeleton-disentangled Representation}. In \bibinfo{booktitle}{\emph{IEEE
  International Conference on Computer Vision, ICCV}}.
\newblock


\bibitem[\protect\citeauthoryear{Teed and Deng}{Teed and Deng}{2021}]%
        {droid}
\bibfield{author}{\bibinfo{person}{Zachary Teed} {and} \bibinfo{person}{Jia
  Deng}.} \bibinfo{year}{2021}\natexlab{}.
\newblock \showarticletitle{{DROID-SLAM: Deep Visual SLAM for Monocular,
  Stereo, and RGB-D Cameras}}.
\newblock \bibinfo{journal}{\emph{Advances in neural information processing
  systems}} (\bibinfo{year}{2021}).
\newblock


\bibitem[\protect\citeauthoryear{Teed, Lipson, and Deng}{Teed
  et~al\mbox{.}}{2024}]%
        {dpvo}
\bibfield{author}{\bibinfo{person}{Zachary Teed}, \bibinfo{person}{Lahav
  Lipson}, {and} \bibinfo{person}{Jia Deng}.} \bibinfo{year}{2024}\natexlab{}.
\newblock \showarticletitle{Deep patch visual odometry}.
\newblock \bibinfo{journal}{\emph{Advances in Neural Information Processing
  Systems}}  \bibinfo{volume}{36} (\bibinfo{year}{2024}).
\newblock


\bibitem[\protect\citeauthoryear{Tevet, Raab, Gordon, Shafir, Cohen-or, and
  Bermano}{Tevet et~al\mbox{.}}{2023}]%
        {mdm}
\bibfield{author}{\bibinfo{person}{Guy Tevet}, \bibinfo{person}{Sigal Raab},
  \bibinfo{person}{Brian Gordon}, \bibinfo{person}{Yoni Shafir},
  \bibinfo{person}{Daniel Cohen-or}, {and} \bibinfo{person}{Amit~Haim
  Bermano}.} \bibinfo{year}{2023}\natexlab{}.
\newblock \showarticletitle{Human Motion Diffusion Model}. In
  \bibinfo{booktitle}{\emph{The Eleventh International Conference on Learning
  Representations}}.
\newblock
\urldef\tempurl%
\url{https://openreview.net/forum?id=SJ1kSyO2jwu}
\showURL{%
\tempurl}


\bibitem[\protect\citeauthoryear{Vaswani, Shazeer, Parmar, Uszkoreit, Jones,
  Gomez, Kaiser, and Polosukhin}{Vaswani et~al\mbox{.}}{2017}]%
        {transformer}
\bibfield{author}{\bibinfo{person}{Ashish Vaswani}, \bibinfo{person}{Noam
  Shazeer}, \bibinfo{person}{Niki Parmar}, \bibinfo{person}{Jakob Uszkoreit},
  \bibinfo{person}{Llion Jones}, \bibinfo{person}{Aidan~N Gomez},
  \bibinfo{person}{\L~ukasz Kaiser}, {and} \bibinfo{person}{Illia Polosukhin}.}
  \bibinfo{year}{2017}\natexlab{}.
\newblock \showarticletitle{Attention is All you Need}. In
  \bibinfo{booktitle}{\emph{Advances in Neural Information Processing
  Systems}}, \bibfield{editor}{\bibinfo{person}{I.~Guyon},
  \bibinfo{person}{U.~Von Luxburg}, \bibinfo{person}{S.~Bengio},
  \bibinfo{person}{H.~Wallach}, \bibinfo{person}{R.~Fergus},
  \bibinfo{person}{S.~Vishwanathan}, {and} \bibinfo{person}{R.~Garnett}}
  (Eds.), Vol.~\bibinfo{volume}{30}. \bibinfo{publisher}{Curran Associates,
  Inc.}
\newblock
\urldef\tempurl%
\url{https://proceedings.neurips.cc/paper_files/paper/2017/file/3f5ee243547dee91fbd053c1c4a845aa-Paper.pdf}
\showURL{%
\tempurl}


\bibitem[\protect\citeauthoryear{von Marcard, Henschel, Black, Rosenhahn, and
  Pons-Moll}{von Marcard et~al\mbox{.}}{2018}]%
        {3dpw}
\bibfield{author}{\bibinfo{person}{Timo von Marcard}, \bibinfo{person}{Roberto
  Henschel}, \bibinfo{person}{Michael Black}, \bibinfo{person}{Bodo Rosenhahn},
  {and} \bibinfo{person}{Gerard Pons-Moll}.} \bibinfo{year}{2018}\natexlab{}.
\newblock \showarticletitle{Recovering Accurate 3D Human Pose in The Wild Using
  IMUs and a Moving Camera}. In \bibinfo{booktitle}{\emph{European Conference
  on Computer Vision (ECCV)}}.
\newblock


\bibitem[\protect\citeauthoryear{Wan, Li, Tian, Liu, Yi, and Li}{Wan
  et~al\mbox{.}}{2021}]%
        {maed}
\bibfield{author}{\bibinfo{person}{Ziniu Wan}, \bibinfo{person}{Zhengjia Li},
  \bibinfo{person}{Maoqing Tian}, \bibinfo{person}{Jianbo Liu},
  \bibinfo{person}{Shuai Yi}, {and} \bibinfo{person}{Hongsheng Li}.}
  \bibinfo{year}{2021}\natexlab{}.
\newblock \showarticletitle{Encoder-decoder with Multi-level Attention for 3D
  Human Shape and Pose Estimation}. In \bibinfo{booktitle}{\emph{The IEEE
  International Conference on Computer Vision (ICCV)}}.
\newblock


\bibitem[\protect\citeauthoryear{Wang and Daniilidis}{Wang and
  Daniilidis}{2023}]%
        {refit}
\bibfield{author}{\bibinfo{person}{Yufu Wang} {and} \bibinfo{person}{Kostas
  Daniilidis}.} \bibinfo{year}{2023}\natexlab{}.
\newblock \showarticletitle{Refit: Recurrent fitting network for 3d human
  recovery}. In \bibinfo{booktitle}{\emph{Proceedings of the IEEE/CVF
  International Conference on Computer Vision}}. \bibinfo{pages}{14644--14654}.
\newblock


\bibitem[\protect\citeauthoryear{Wang, Wang, Liu, and Daniilidis}{Wang
  et~al\mbox{.}}{2024}]%
        {tram}
\bibfield{author}{\bibinfo{person}{Yufu Wang}, \bibinfo{person}{Ziyun Wang},
  \bibinfo{person}{Lingjie Liu}, {and} \bibinfo{person}{Kostas Daniilidis}.}
  \bibinfo{year}{2024}\natexlab{}.
\newblock \showarticletitle{TRAM: Global Trajectory and Motion of 3D Humans
  from in-the-wild Videos}.
\newblock \bibinfo{journal}{\emph{arXiv preprint arXiv:2403.17346}}
  (\bibinfo{year}{2024}).
\newblock


\bibitem[\protect\citeauthoryear{Wei, Lin, Liu, and Liao}{Wei
  et~al\mbox{.}}{2022}]%
        {mpsnet}
\bibfield{author}{\bibinfo{person}{Wen-Li Wei}, \bibinfo{person}{Jen-Chun Lin},
  \bibinfo{person}{Tyng-Luh Liu}, {and} \bibinfo{person}{Hong-Yuan~Mark Liao}.}
  \bibinfo{year}{2022}\natexlab{}.
\newblock \showarticletitle{Capturing Humans in Motion: Temporal-Attentive 3D
  Human Pose and Shape Estimation from Monocular Video}. In
  \bibinfo{booktitle}{\emph{The IEEE Conference on Computer Vision and Pattern
  Recognition (CVPR)}}.
\newblock


\bibitem[\protect\citeauthoryear{Xu, Zhang, Zhang, and Tao}{Xu
  et~al\mbox{.}}{2022}]%
        {vitpose}
\bibfield{author}{\bibinfo{person}{Yufei Xu}, \bibinfo{person}{Jing Zhang},
  \bibinfo{person}{Qiming Zhang}, {and} \bibinfo{person}{Dacheng Tao}.}
  \bibinfo{year}{2022}\natexlab{}.
\newblock \showarticletitle{Vi{TP}ose: Simple Vision Transformer Baselines for
  Human Pose Estimation}. In \bibinfo{booktitle}{\emph{Advances in Neural
  Information Processing Systems}}.
\newblock


\bibitem[\protect\citeauthoryear{Ye, Pavlakos, Malik, and Kanazawa}{Ye
  et~al\mbox{.}}{2023}]%
        {slahmr}
\bibfield{author}{\bibinfo{person}{Vickie Ye}, \bibinfo{person}{Georgios
  Pavlakos}, \bibinfo{person}{Jitendra Malik}, {and} \bibinfo{person}{Angjoo
  Kanazawa}.} \bibinfo{year}{2023}\natexlab{}.
\newblock \showarticletitle{Decoupling Human and Camera Motion from Videos in
  the Wild}. In \bibinfo{booktitle}{\emph{IEEE Conference on Computer Vision
  and Pattern Recognition (CVPR)}}.
\newblock


\bibitem[\protect\citeauthoryear{Yi, Zhou, and Xu}{Yi et~al\mbox{.}}{2021}]%
        {transpose}
\bibfield{author}{\bibinfo{person}{Xinyu Yi}, \bibinfo{person}{Yuxiao Zhou},
  {and} \bibinfo{person}{Feng Xu}.} \bibinfo{year}{2021}\natexlab{}.
\newblock \showarticletitle{TransPose: Real-time 3D Human Translation and Pose
  Estimation with Six Inertial Sensors}.
\newblock \bibinfo{journal}{\emph{ACM Transactions on Graphics}}
  \bibinfo{volume}{40}, \bibinfo{number}{4}, Article \bibinfo{articleno}{86}
  (\bibinfo{date}{08} \bibinfo{year}{2021}).
\newblock


\bibitem[\protect\citeauthoryear{Yin, Cai, Wang, Wang, Wei, Mei, Xiao, Yang,
  Sun, Yamashita, et~al\mbox{.}}{Yin et~al\mbox{.}}{2024}]%
        {whac}
\bibfield{author}{\bibinfo{person}{Wanqi Yin}, \bibinfo{person}{Zhongang Cai},
  \bibinfo{person}{Ruisi Wang}, \bibinfo{person}{Fanzhou Wang},
  \bibinfo{person}{Chen Wei}, \bibinfo{person}{Haiyi Mei},
  \bibinfo{person}{Weiye Xiao}, \bibinfo{person}{Zhitao Yang},
  \bibinfo{person}{Qingping Sun}, \bibinfo{person}{Atsushi Yamashita},
  {et~al\mbox{.}}} \bibinfo{year}{2024}\natexlab{}.
\newblock \showarticletitle{WHAC: World-grounded Humans and Cameras}.
\newblock \bibinfo{journal}{\emph{arXiv preprint arXiv:2403.12959}}
  (\bibinfo{year}{2024}).
\newblock


\bibitem[\protect\citeauthoryear{Yu, Park, and Lee}{Yu et~al\mbox{.}}{2021}]%
        {humandynamic}
\bibfield{author}{\bibinfo{person}{Ri Yu}, \bibinfo{person}{Hwangpil Park},
  {and} \bibinfo{person}{Jehee Lee}.} \bibinfo{year}{2021}\natexlab{}.
\newblock \showarticletitle{Human dynamics from monocular video with dynamic
  camera movements}.
\newblock \bibinfo{journal}{\emph{ACM Trans. Graph.}} \bibinfo{volume}{40},
  \bibinfo{number}{6}, Article \bibinfo{articleno}{208} (\bibinfo{date}{dec}
  \bibinfo{year}{2021}), \bibinfo{numpages}{14}~pages.
\newblock
\showISSN{0730-0301}
\urldef\tempurl%
\url{https://doi.org/10.1145/3478513.3480504}
\showDOI{\tempurl}


\bibitem[\protect\citeauthoryear{Yuan, Iqbal, Molchanov, Kitani, and
  Kautz}{Yuan et~al\mbox{.}}{2022}]%
        {glamr}
\bibfield{author}{\bibinfo{person}{Ye Yuan}, \bibinfo{person}{Umar Iqbal},
  \bibinfo{person}{Pavlo Molchanov}, \bibinfo{person}{Kris Kitani}, {and}
  \bibinfo{person}{Jan Kautz}.} \bibinfo{year}{2022}\natexlab{}.
\newblock \showarticletitle{GLAMR: Global Occlusion-Aware Human Mesh Recovery
  with Dynamic Cameras}. In \bibinfo{booktitle}{\emph{Proceedings of the
  IEEE/CVF Conference on Computer Vision and Pattern Recognition (CVPR)}}.
\newblock


\bibitem[\protect\citeauthoryear{Zhang, Tian, Zhang, Li, An, Sun, and
  Liu}{Zhang et~al\mbox{.}}{2023}]%
        {pymafx}
\bibfield{author}{\bibinfo{person}{Hongwen Zhang}, \bibinfo{person}{Yating
  Tian}, \bibinfo{person}{Yuxiang Zhang}, \bibinfo{person}{Mengcheng Li},
  \bibinfo{person}{Liang An}, \bibinfo{person}{Zhenan Sun}, {and}
  \bibinfo{person}{Yebin Liu}.} \bibinfo{year}{2023}\natexlab{}.
\newblock \showarticletitle{PyMAF-X: Towards Well-aligned Full-body Model
  Regression from Monocular Images}.
\newblock \bibinfo{journal}{\emph{IEEE Transactions on Pattern Analysis and
  Machine Intelligence}} (\bibinfo{year}{2023}).
\newblock


\end{thebibliography}

\end{document}